\documentclass[10pt]{article}
\usepackage[utf8]{inputenc}
\usepackage{amsmath, amssymb, amsthm}
\usepackage{graphicx}
\usepackage{hyperref}
\usepackage{geometry}
\usepackage{times}  
\usepackage{natbib}  
\usepackage{authblk}  
\usepackage{anyfontsize} 

\usepackage{array}
\usepackage{tabu}
\usepackage{wrapfig}
\usepackage{float}
\usepackage{thm-restate}
\usepackage{graphicx}
\usepackage{subcaption}
\usepackage{hyperref}
\usepackage{url}
\usepackage{xcolor}
\usepackage{multirow}
\usepackage{booktabs}
\usepackage[ruled,vlined]{algorithm2e}
\usepackage{bm}

\newcommand{\EDremove}[1]{{}}

\usepackage{amsmath,amsfonts,bm}
\usepackage{amstext}
\usepackage{amssymb}
\usepackage{bm}
\usepackage{enumitem}

\def\eqref#1{equation~\ref{#1}}

\def\1{\bm{1}}

\DeclareMathAlphabet{\mathsfit}{\encodingdefault}{\sfdefault}{m}{sl}
\SetMathAlphabet{\mathsfit}{bold}{\encodingdefault}{\sfdefault}{bx}{n}

\makeatletter
\newcommand*{\@opargbegintheorem}[3]{\trivlist
      \item[\hskip \labelsep{\bfseries #1\ #2}] \textbf{(#3)}\ \itshape}
\makeatother

\newtheorem{theorem}{Theorem}
\newtheorem{lemma}{Lemma} 
\newtheorem{proposition}{Proposition}

\newtheorem{definition}{Definition}

\newtheorem{assumption}{Assumption}

\author[1,4]{Yudong Luo}
\author[2]{Yangchen Pan}
\author[3]{Jiaqi Tan}
\author[1,4]{Pascal Poupart}

\affil[1]{School of Computer Science, University of Waterloo}
\affil[2]{Department of Engineering Science, University of Oxford}
\affil[3]{School of Computing Science, Simon Fraser University}
\affil[4]{Vector Institute}

\title{Measures of Variability for Risk-averse Policy Gradient}

\date{}

\begin{document}
\maketitle

\begin{abstract} \normalsize
Risk-averse reinforcement learning (RARL) is critical for decision-making under uncertainty, which is especially valuable in high-stake applications. However, most existing works focus on risk measures, e.g., conditional value-at-risk (CVaR), while measures of variability remain underexplored. In this paper, we comprehensively study nine common measures of variability, namely Variance, Gini Deviation, Mean Deviation, Mean-Median Deviation, Standard Deviation, Inter-Quantile Range, CVaR Deviation, Semi\_Variance, and Semi\_Standard Deviation. Among them, four metrics have not been previously studied in RARL. We derive policy gradient formulas for these unstudied metrics, improve gradient estimation for Gini Deviation, analyze their gradient properties, and incorporate them with the REINFORCE and PPO frameworks to penalize the dispersion of returns.

Our empirical study reveals that variance-based metrics lead to unstable policy updates. In contrast, CVaR Deviation and Gini Deviation show consistent performance across different randomness and evaluation domains, achieving high returns while effectively learning risk-averse policies. Mean Deviation and Semi\_Standard Deviation are also competitive across different scenarios. This work provides a comprehensive overview of variability measures in RARL, offering practical insights for risk-aware decision-making and guiding future research on risk metrics and RARL algorithms.
\end{abstract}


\section{Introduction}
The demand for avoiding risks in practical applications has inspired risk-averse reinforcement learning (RARL). For example, we want to avoid collisions in autonomous driving~\citep{naghshvar2018risk}, or avoid huge financial losses in portfolio management~\citep{bjork2014mean}. In addition to conventional RL, which finds policies to maximize expected return~\citep{sutton2018reinforcement}, RARL also considers the control of risk.

Many risk metrics have been studied for RARL, with a focus on ``risk measures", for instance,  value-at-risk (VaR)~\citep{chow2017risk}, conditional value-at-risk (CVaR)~\citep{bauerle2011markov,chow2014algorithms,chow2015risk,greenberg2022efficient,luo2024simple}, entropic risk measure~\citep{fei2021risk,hau2023entropic}. In this paper, however, we focus on a different class of risk metrics known as ``measures of variability". Although both risk measures and measures of variability quantify the risk (or uncertainty) of random variables, they adhere to different sets of axioms. For example, while translation invariance is commonly expected for risk measures, it is not required for measures of variability. Instead, location invariance is a defining property of measures of variability (see Sec.~\ref{sec:measure_var} for further discussion). Despite these differences, the two types of metrics are interrelated and can sometimes be transformed into one another. For instance, \cite{tamar2015policy} demonstrated that the Semi\_Standard Deviation plus the mean constitutes a coherent risk measure, while \cite{rockafellar2006generalized} showed that variability is closely linked to strictly expectation-bounded risk measures.


Measures of variability generally describe the dispersion of a random variable, with variance being the most commonly used metric. Such a paradigm in RL is usually referred to as mean-variance RL, which aims to maximize the expected total return while minimizing its variance~\citep{tamar2012policy,la2013actor,xie2018block,bisi2020risk,zhang2021mean}. Variance related metrics, e.g., Standard Deviation~\citep{yang2021wcsac}, Semi\_Standard Deviation~\citep{tamar2015policy}, and Semi\_Variance~\citep{ma2022mean} have also been explored in RL. However, despite the importance of measures of variability as a class of risk metrics, there is currently no comprehensive study in RL that explores metrics beyond variance, nor any systematic comparison of their effectiveness in finding risk-averse policies by penalizing return dispersion.

In this paper, we present the first comprehensive study of RARL using a diverse set of measures of variability beyond Variance. While Variance has been widely adopted, we systematically investigate eight alternative metrics, namely Gini Deviation, Mean Deviation, Mean-Median Deviation, Standard Deviation, Inter-Quantile Range, CVaR Deviation, Semi\_Variance, and Semi\_Standard Deviation, derive their policy gradient formulas, and when applicable, provide error bounds for biased estimators. Notably, four of them (Gini Deviation, Mean Deviation, Mean-Median Deviation, and Semi\_Standard Deviation) have not been formally studied in RARL prior to our work. Our empirical analysis addresses critical open questions: (1) whether these measures of variability effectively encourage risk-averse policies, (2) whether such policies sacrifice expected return, and (3) how robust these metrics are across different sources of randomness and domains. A key finding is that several mathematically distinct metrics, such as CVaR Deviation and Gini Deviation, exhibit consistent performance. Before our work, it was unclear whether these metrics would lead to similar risk-averse behaviors in practice. Additionally, we identify previously overlooked measures, such as Mean Deviation and Semi\_Standard Deviation, that perform competitively with established approaches, offering practical alternatives for RARL. Our results provide the first systematic comparison of these variability metrics, offering actionable insights for practitioners and advancing the understanding of risk-averse policy optimization.

\section{Preliminaries}


\subsection{Risk Measures and Measures of Variability}
\label{sec:measure_var}

Risk arises from the uncertainty of a random variable, and various metrics are defined to quantify this uncertainty (and therefore risk). Among them, \textit{risk measures} and \textit{measures of variability} are two commonly studied categories. (Note: The terms ``risk measures" and ``measures of variability"  are commonly used to refer to concepts with distinct sets of desired properties. Here, we use the term ``risk metrics" as a more inclusive, general category that encompasses all of these risk-related measures.)

Different \textit{risk measures} usually convey different meanings. For example, CVaR represents the expected value of the distribution's tail, while VaR is analogous to a quantile. On the other hand, \textit{measures of variability} (also named deviation measures) aim to capture the dispersion (i.e., how spread out) of a random variable. We use the term ``measures of variability" in accordance with \cite{furman2017gini}. Importantly, they are not ``risk measures" in the sense defined by \cite{artzner1999coherent}. We briefly differentiate them below.


When describing general random variables, we work with an atomless probability space $(\Omega, \mathcal{F}, \mathbb{P})$. Let $\mathcal{M}$ denote the set of real random variables. Let $\mathcal{M}^p$ denote the set of random variables with finite $p$-th moment, $p\in[0,\infty)$, and let $\mathcal{M}^\infty$ be the set of all essentially bounded random variables. For every random variable $X\in \mathcal{M}^0$, we use $F_X$ to denote the cdf of X. The inverse cdf is thus $F^{-1}_X$. Given a confidence level $\alpha\in(0,1)$, the $\alpha$-level quantile of $X$ is denoted as $q_\alpha(X)$, and given by
\vspace{-3pt}
\begin{equation}
q_\alpha(X)=\mathrm{inf} \big\{x:\mathrm{Pr}(X \leq x)\geq \alpha\big\}.
\end{equation}
\vspace{-2pt}
Two random variables $X$ and $Y$ have the same distribution under $\mathbb{P}$ is denoted by $X\overset{d}{=} Y$. Let $\mathcal{X}$ denote the convex cones of random variables, of which $\mathcal{M}^\infty$ is always contained. We use $\mathbb{I}$ for the indicator function.

\subsubsection{Risk Measures} A risk measure $\rho$ maps $\mathcal{X}$ to $(-\infty, \infty]$. The following are several properties that are important in the literature on risk measures. Law-invariance is
the property that is satisfied by all the risk metrics we discussed here. 

\noindent (A) Law-invariance: if $X\overset{d}{=}Y$, then $\rho(X)=\rho(Y)$.

The following properties have been standard in the theory of coherent risk measures \citep{artzner1999coherent}:

\noindent (A1) Positive homogeneity: $\rho(c X) = c \rho(X)$ for all $c>0$ and $X\in\mathcal{X}$.

\noindent (A2) Sub-additivity: $\rho(X+Y)\leq \rho(X) + \rho(Y)$ for all $X,Y \in\mathcal{X}$.


\noindent (B1) Monotonicity: $\rho(X)\leq \rho(Y)$ if $X,Y\in \mathcal{X}$ and $X\leq Y$ $\mathbb{P}$-almost surely.

\noindent (B2) Translation invariance: $\rho(X-c)=\rho(X)-c$ for all $c\in \mathbb{R}$ and $X\in\mathcal{X}$.

We refer readers to \cite{delbaen2012monetary} for more explanations of these properties. 

\begin{definition}[\cite{artzner1999coherent}]
    A risk measure is \textbf{coherent} if it satisfies (A1), (A2), (B1) and (B2). 
\end{definition}

If a risk measure satisfies (B1), (B2), and convexity ($\rho(c X + (1-c)Y)\leq c \rho(X) + (1-c)\rho(Y)$ for all $c\in[0,1]$), it is a convex risk measure. If a convex risk measure also satisfies positive homogeneity, then it is a coherent risk measure (since any pair among (A1), (A2) and convexity implies the remaining one). 

\subsubsection{Measures of Variability}
A measure of variability $\nu$ maps $\mathcal{X}$ to $[0,\infty]$. The desirable properties of variability are different from those of risk measures. For instance, we expect a measure of variability to be standardized and location invariant.

\noindent (C1) Standardization: $\nu(m)=0$ for all $m\in\mathbb{R}$.

\noindent (C2) Location invariance: $\nu(X-m)=\nu(X)$ for all $m\in \mathbb{R}$ and $X\in\mathcal{X}$.

At this point, we can see measures of variability are different from risk measures since a function can not satisfy (B2) and (C2) at the same time. Note that currently there is no common agreement on the properties a measure of variability should have. \cite{rockafellar2006generalized} defined ``general deviation measures", but the variance is not included in their definitions. Here we follow the definitions given by \cite{furman2017gini}, which are similar to the axioms proposed by \cite{rockafellar2006generalized} but are easier to interpret.

\begin{definition}[\cite{furman2017gini}]
    A function $\nu:\mathcal{X}\rightarrow [0,\infty]$ is a measure of variability if it satisfies (A), (C1) and (C2). A measure of variability is \textbf{coherent} if it further satisfies (A1) and (A2).
\end{definition}

We mention several examples of measures of variability by following the seminal works of \cite{david1998early,rockafellar2006generalized}.
\begin{itemize}
    \item Variance. $\mathbb{V}[X] = \mathbb{E}[(X-\mathbb{E}[X])^2]$, $X\in \mathcal{M}^2$. It has an alternative expression $\mathbb{V}[X]=\frac{1}{2}\mathbb{E}[(X-X^{*})^2]$ as noted by Corrado Gini, where $X^*$ is an i.i.d copy of $X$. Due to the quadratic function, it distorts the values of $X-X^{*}$ by making them larger when they are outside the interval $[-1,1]$ and smaller otherwise. It is not a coherent measure of variability as it violates (A1). It is widely used in portfolio management given its simplicity and easy interpretation~\citep{markowitz2008portfolio,bjork2014mean}.
    \item Gini Deviation (or Gini Mean Difference). $\mathrm{GD}[X] = \frac{1}{2}\mathbb{E}[|X - X^{*}|]$, $X\in \mathcal{M}^1$, where $X^*$ is an i.i.d copy of $X$. The idea of this metric originates from the work of~\cite{gini1912variabilita} that the representation of variance could be misleading in the sense that the variability of any random variable should not be based on the center of the underlying distribution. It is a coherent measure of variability~\citep{furman2017gini}. It also appears in portfolio management as a replacement for variance since it allows the derivation of necessary conditions for stochastic dominance~\citep{yitzhaki1982stochastic}.
    \item Mean Absolute Deviation from the Mean (or Mean Deviation). $\mathrm{MD}[X] = \mathbb{E}[|X-\mathbb{E}[X]|]$, $X\in \mathcal{M}^1$. It is a coherent measure of variability (see Appendix~\ref{sec:md-coherent}). It is also used in portfolio management since the problem can be reduced to a linear programming problem, instead
    of a quadratic programming problem in the case of variance~\citep{konno1999mean}.
    \item Mean Absolute Deviation from the Median (or Mean-Median Deviation). $\mathrm{MMD}[X] = \mathbb{E}[|X-\mathrm{Median}(X)|] = \min_{x\in\mathbb{R}}\mathbb{E}[|X-x|]$, $X\in \mathcal{M}^1$. The second equality is due to the fact that the median is the minimum value of the L1 estimate. It is a coherent measure of variability (see Appendix~\ref{sec:mmd-coherent}). It is an alternative to Mean Deviation, since the median is more robust to outliers and skewed distributions.
    \item Standard Deviation. $\mathrm{STD}[X] = \sqrt{\mathbb{V}[X]}$. The square root does not rectify the distortion mentioned in variance, since the expectation is inside the square root. It is a coherent measure of variability~\citep{furman2017gini}. 
\end{itemize}

There are other measures of variability using partial information of a random variable to compute variability.
\begin{itemize}
    \item Inter-Quantile Range. $\mathrm{IQR}_\alpha[X] = F_X^{-1}(\alpha) - F^{-1}_X(1-\alpha), \alpha\in[\frac{1}{2},1)$, $X\in \mathcal{M}^0$. It is not a coherent measure of variability as it violates (A2) (since quantile fails to be subadditive). It plays a key role in the construction of a box plot~\citep{spitzer2014boxplotr}. 
    \item CVaR Deviation. It can be defined on either the lower or upper tail of $X$. To make sure the deviation is non-negative, we distinguish the lower tail CVaR (denoted as $\mathrm{CVaR}^{\lor}$) and upper tail CVaR (denoted as $\mathrm{CVaR}^{\land}$). The upper tail CVaR at confidence level $\alpha$ is given by $\mathrm{CVaR}^\land_\alpha(X)=\frac{1}{1-\alpha}\int_\alpha^1 q_\beta(X) d\beta$, where $q_\beta(X)$ is the quantile. In this case, CVaR deviation is $\mathrm{CD}[X]=\mathrm{CVaR}_\alpha^\land(X-\mathbb{E}[X])=\mathrm{CVaR}_\alpha^\land(X)-\mathbb{E}[X], X\in \mathcal{M}^{1}$ (since CVaR is translation invariant). The lower tail CVaR at confidence level $\alpha$ is given by $\mathrm{CVaR}^\lor_\alpha(X)=\frac{1}{\alpha}\int^\alpha_0 q_\beta(X) d\beta$. In this case, CVaR deviation is $\mathrm{CD}[X]=\mathbb{E}[X]-\mathrm{CVaR}^\lor_\alpha(X)$. $\mathrm{CVaR}^\land_\alpha(X - \mathbb{E}[X])$ is a coherent measure of variability (see Appendix~\ref{sec:cd-coherent}). We see that CVaR Deviation equals a mean-CVaR problem. Mean-CVaR criteria is widely used in portfolio management~\citep{yao2013mean} and also RL~\citep{ijcai2022p0510} to avoid potential losses.
    \item Semi\_Variance and Semi\_Standard Deviation. The definition of these two can use either the upside or downside of a distribution. Here we give examples using the downside. Note that in the literature, some define Semi\_Variance as a conditional expectation, i.e., $\mathbb{E}[(X-\mathbb{E}[X])^2 | {X\leq \mathbb{E}[X]}]$. While it is more often used as defining with an indicator, i.e., $\mathrm{S}\mathbb{V}[X]=\mathbb{E}[(X-\mathbb{E}[X])^2\mathbb{I}_{X\leq \mathbb{E}[X]}], X\in \mathcal{M}^2$. In this paper, we adopt the latter definition for easier optimization. Semi\_Standard Deviation (or simply Semi\_Deviation) is therefore defined as $\mathrm{SD}[X]=\sqrt{\mathrm{S}\mathbb{V}[X]}$. Semi\_Variance is not coherent as it violates (A1). Semi\_Deviation defined as $\sqrt{\mathrm{S}\mathbb{V}[X]}$ is a coherent measure of variability (see Appendix~\ref{sec:sstd-coherent}). Semi\_Variance is widely used to measure the downside risk in portfolio management since it is consistent with the intuitive perception of risk of investors~\citep{boasson2011portfolio}.
\end{itemize}

\textit{Remark.} 
The aforementioned metrics are mainly used in finance. In RL, apart from variance, Gini Deviation is discussed in our previous paper, but with a biased policy gradient estimation ~\citep{luo2023alternative}. Standard Deviation is explored by \cite{yang2021wcsac,kim2022trc}. These two originally aim to optimize the CVaR of the return. By assuming the return distribution is Gaussian, CVaR optimization simplifies to a mean-STD problem. The CVaR policy gradient (part of the CVaR Deviation gradient) is provided by \cite{tamar2015optimizing}. Semi\_Variance is studied by \citet{ma2022mean} in the average reward setting, unlike the non-average setting considered in this paper. The mean-Semi\_Deviation policy gradient is derived by \cite{tamar2015policy}. We are not aware of other RL techniques that incorporate the remaining measures of variability.

\subsubsection{Signed Choquet Integral}

We also introduce the concept of signed Choquet integral, which will be later used in Sec.~\ref{sec:other-measures} for gradient calculation. It provides alternative definitions for some metrics (e.g., Gini Deviation, Mean-Median Deviation) and makes gradient-based optimization convenient.

The Choquet integral~\citep{choquet1954theory} was first used in statistical mechanics and potential theory and was later applied to decision making as a way of measuring the expected utility~\citep{grabisch1996application}. The signed Choquet integral belongs to the Choquet integral family and is defined as

\begin{definition}[\cite{wang2020characterization}]
\label{def:choquet}
A signed Choquet integral $\Phi_h: X \rightarrow \mathbb{R}$ is defined as
\begin{equation}
    \Phi_h(X)=\int^0_{-\infty}\Big(h\big(\mathrm{Pr}(X\geq x)\big)-h(1)\Big)dx+\int^\infty_0 h\big(\mathrm{Pr}(X\geq x)\big)dx,
\end{equation}
\noindent where $h$ is the distortion function and $h\in\mathcal{H}$ such that $\mathcal{H}=\{h:[0, 1]\rightarrow\mathbb{R}, h(0)=0, h~ \mathrm{is~of~bounded~variation}\}$.
\end{definition}
This integral has become the building block of law-invariant risk metrics after the work of \citet{kusuoka2001law, grechuk2009maximum}. One reason why the signed Choquet integral is of interest to the risk research community is that it is not necessarily monotone (axiom (B1)). Since most practical measures of variability are not monotone, e.g., variance, standard deviation, or deviation measures in~\citet{rockafellar2006generalized}, it is possible to represent these measures in terms of $\Phi_h$ by choosing a specific distortion function $h$.

The signed Choquet integral has an alternative representation by using its quantile under some mild assumption, which is stated in Lemma ~\ref{le:quantile_rep}.

\begin{lemma}[\cite{wang2020characterization}, Lemma 3]
\label{le:quantile_rep}
$\Phi_h(X)$ has a quantile representation. If $F^{-1}_X$ is continuous, then $\Phi_h(X)=\int^1_0 F^{-1}_X(1-\alpha)dh(\alpha)$, where $F^{-1}_X$ is the inverse cdf (quantile function) of X.
\end{lemma}

\subsection{RL Background and Policy Gradient}
\label{sec:pg_background}
We also cover the essential background on policy gradient in RL. In standard RL settings, the agent-environment interaction is modeled as a Markov decision process (MDP), represented as a tuple $(\mathcal{S}, \mathcal{A}, R, P, \mu_0, \gamma)$~\citep{puterman2014markov}. $\mathcal{S}$ and $\mathcal{A}$ denote state and action spaces. $P(\cdot|s,a)$ defines the transition. $R$ is the state and action dependent reward variable, $\mu_0$ is the initial state distribution, and $\gamma\in(0, 1]$ is the discount factor. We use $r_t$ or $r(s, a)$ to represent the reward that the agent actually gets at time step $t$ or when it visits state $s$ and takes action $a$. We may also overload the notation of $P$ to represent the probability of other quantities when the context is clear. 

An agent follows its policy $\pi: \mathcal{S} \times \mathcal{A} \rightarrow [0, +\infty)$ in the environment. Its return at time step $t$ is defined as (denote $R_{t+1}=R(S_t,A_t)$)
\begin{equation}
\label{eq:G_t}
\begin{aligned}
     G_t\overset{\mathrm{def}}{=} &R_{t+1} + \gamma R_{t+1} + \gamma^2 R_{t+3} + ...\\
     =& R_{t+1} + \gamma G_{t+1}~.
\end{aligned}
\end{equation}
Thus, $G_0$ is the random variable that indicates the total return starting from the initial state following $\pi$. It is usually convenient to define the value function to represent the expected return the agent can get when it is in some status. The state value function is defined as $V^\pi(s) \overset{\mathrm{def}}{=} \mathbb{E}[G_t|S_t=s, \pi]$.  The state-action value function is defined as $Q^\pi(s,a) \overset{\mathrm{def}}{=} \mathbb{E}[G_t|S_t=s, A_t=a, \pi]$.

Traditional (risk-neutral) RL aims to maximize the expectation of $G_0$. A common approach is to apply the time difference learning strategy to maximize $Q(s,a)$ based on transitions. Here we introduce another line of approach named the policy gradient. In risk-averse setting, since most risk metrics are nonlinear and lack the monotonicity of dynamic programming, improving the policy via policy gradient is desired. We describe the policy gradient in risk-neural setting, and the gradient in risk-averse setting bears the similarity.


Suppose that the policy $\pi$ is parameterized by $\theta$, denoted by $\pi_\theta$. Policy gradient methods directly take the gradient for $\mathbb{E}[G_0]$, i.e., calculating $\nabla_\theta \mathbb{E}[G_0]$, then perform the gradient ascent. We derive the gradient via Monte Carlo sampling. By executing policy $\pi_\theta$ in the environment, we can collect a trajectory $\tau=(s_0,a_0,r_1,s_1,a_1,...,r_T,s_T)$, with the return of this trajectory being $R_\tau=r_1+\gamma r_2 +...+\gamma^{T-1}r_T$. Note that $\mathbb{E}[G_0]=\mathbb{E}_\tau[R_\tau]$, thus $\nabla_\theta \mathbb{E}[G_0]=\nabla_\theta \mathbb{E}_\tau[R_\tau]$, and

\begin{equation}
\label{eq:mc_pg}
\begin{aligned}
    \nabla_\theta \mathbb{E}_\tau[R_\tau]&=\nabla_\theta \sum_\tau P(\tau;\theta) R_\tau = \sum_\tau R_\tau P(\tau;\theta)\frac{\nabla_\theta P(\tau;\theta)}{P(\tau;\theta)}\\
    &=\sum_\tau R_\tau P(\tau;\theta) \nabla_\theta\log P(\tau;\theta) = \mathbb{E}_\tau [R_\tau \nabla_\theta\log P(\tau;\theta)],
\end{aligned}
\end{equation}
where we applied the equality $\nabla_\theta \log(x)=\frac{1}{x}\nabla_\theta x$ to the second line. The probability of a trajectory $P(\tau;\theta)$ is defined as $P(\tau;\theta) = \mu(s_0) \prod_{t=0}^{T-1}\Big[\pi_\theta(a_t|s_t) P(s_{t+1}|s_t,a_t)\Big]$. Thus, the logarithm is
$\log P(\tau;\theta) = \log \mu(s_0) + \sum_{t=0}^{T-1}\Big[\log\pi_\theta(a_t|s_t) + \log P(s_{t+1}|s_t,a_t)\Big]$, whose gradient is
\begin{equation}
\label{eq:sum_log_pi}
    \nabla_\theta\log P(\tau;\theta)=\nabla_\theta \sum_{t=0}^{T-1} \log\pi_\theta(a_t|s_t).
\end{equation}
Combining Eq.~\ref{eq:mc_pg} and~\ref{eq:sum_log_pi}, the policy gradient is
\begin{equation}
\label{eq:naive_pg}
    \nabla_\theta \mathbb{E}_\tau[R_\tau]=\mathbb{E}_\tau \Big[R_\tau  \sum_{t=0}^{T-1}\nabla_\theta\log\pi_\theta(a_t|s_t)\Big].
\end{equation}
The gradient in Eq.~\ref{eq:naive_pg} is known to have high variance. The variance can be reduced by the following two commonly used techniques. First, there exist cross time terms in Eq.~\ref{eq:naive_pg}, i.e., $r_i \nabla_\theta \log \pi_\theta(a_j|s_j)$ with $i<j$, that can be removed due to zero expectation. Denote the return starting from time $t$ (or named as reward-to-go) of trajectory $\tau$ as $R_{\tau,t}$ ($R_{\tau,t}\overset{\mathrm{def}}{=} \sum_{t'=t+1}^{T}\gamma^{t'-t-1} r_{t'}$). Removing those cross time terms yields
\begin{equation}
\label{eq:reinforce}
    \nabla_\theta \mathbb{E}_\tau[R_\tau] = \mathbb{E}_{\tau}\Big[\sum_{t=0}^{T-1} \nabla_\theta \log\pi_\theta(a_t|s_t) \gamma^t R_{\tau,t}\Big].
\end{equation}
Second, the term $R_{\tau,t}$ in Eq.~\ref{eq:reinforce} can be regarded as the Monte Carlo estimation of $Q^\pi(s_t,a_t)$. Based on the concept of control variate, we can subtract $V^\pi(s_t)$ from it to reduce variance, since $V^\pi(s_t)=\mathbb{E}_{a_t}Q^\pi(s_t,a_t)$, which results in
\begin{equation}
\label{eq:rf_bl}
    \nabla_\theta \mathbb{E}_\tau[R_\tau] = \mathbb{E}_{\tau}\Big[\sum_{t=0}^{T-1} \nabla_\theta \log\pi_\theta(a_t|s_t) \gamma^t \big(R_{\tau,t}-V^\pi(s_t)\big)\Big].
\end{equation}
Eq.~\ref{eq:rf_bl} is more often known as REINFORCE with baseline, where $V^\pi(s_t)$ is the baseline. In practice, we need to estimate this baseline using another function.

Given $n$ sampled trajectories $\{\tau_i\}_{i=1}^n$ under policy $\pi_\theta$, the unbiased estimator for Eq.~\ref{eq:rf_bl} is therefore
\vspace{-2pt}
\begin{equation}
\label{eq:rf_est}
    \nabla_\theta \mathbb{E}_\tau[R_\tau] = \frac{1}{n} \sum_{i=1}^n \sum_{t=0}^{T-1} \nabla_\theta \log\pi_\theta(a_{i,t}|s_{i,t}) \gamma^t \big(R_{{\tau_i},t}-V^\pi(s_{i,t})\big).
\end{equation}
\vspace{-5pt}

\section{Policy Gradient Derivation for Measures of Variability}
\label{sec:other-measures}

In this section, we discuss optimizing measures of variability of total return $G_0$ in RL with a gradient-based approach using the metrics described in Sec.~\ref{sec:measure_var}. For each metric, we first derive its gradient for a general random variable $X$, then extend it to the policy gradient.

For random variable $X$, we assume its distribution function is parameterized by $\theta$. In RL, we can interpret $\theta$ as the policy parameters and $X$ as the return variable under that policy, i.e., $G_0$. Denote the pdf of $X$ as $f_X(x;\theta)$, and the $\alpha$-quantile as $q_\alpha(X;\theta)$. For technical convenience, we make the following assumptions, which are also realistic in RL. 

\begin{assumption}
\label{assump:1}
$X$ is a continuous random variable bounded in range $[-b, b]$, with a continuous and bounded density function $f_X(x;\theta)$ for all $\theta$.
\end{assumption}
\begin{assumption}
\label{assump:2}
$\frac{\partial}{\partial \theta_i}q_\alpha(X;\theta)$ exists and is bounded for all $\theta$, where $\theta_i$ is the $i$-th element of $\theta$.
\end{assumption}
\begin{assumption}
\label{assump:3}
$\frac{\partial f_X(x;\theta)}{\partial \theta_i}/f_X(x;\theta)$ exists and is bounded for all $\theta,x$, where $\theta_i$ is the $i$-th element of $\theta$.
\end{assumption}
Since $X$ is continuous, $q_\alpha(X;\theta) = F^{-1}_X(\alpha)$. Also, the second assumption is satisfied whenever $\frac{\partial}{\partial\theta_i}f_X(x;\theta)$ is bounded. These assumptions are common in likelihood-ratio methods~\citep{tamar2015optimizing}. Relaxing them is possible but would complicate the presentation.

\subsection{Variance}
\label{sec:mean-variance-rl}
Variance is widely used in  RARL. This line of research is also called mean-variance RL. Note that in the literature, there are two ways to define a variance-based risk. The first one defines the variance based on the Monte Carlo \textbf{total return} $G_0$~\citep{tamar2012policy,la2013actor,xie2018block}. The second defines the variance on the \textbf{per-step reward} $R$, which seeks to optimize an upper bound of $\mathbb{V}[G_0]$~\citep{bisi2020risk,zhang2021mean}. In this paper, we mainly focus on a direct measurement of the variability of $G_0$. Interested readers may refer to  Appendix~\ref{sec:per-step-r-var} for a discussion of $\mathbb{V}[R]$-based methods. 

\subsubsection{Mean-Variance Policy Gradient}

For a random variable $X$, variance is defined as
\begin{equation}
    \mathbb{V}[X]=\mathbb{E}[(X-\mathbb{E}[X])^2]= \mathbb{E}[X^2]-(\mathbb{E}[X])^2,
\end{equation}
whose gradient is therefore
\begin{equation}
\label{eq:variance_grad}
    \nabla_\theta \mathbb{V}[X]=\nabla_\theta \mathbb{E}[X^2]-2\mathbb{E}[X]\nabla_\theta\mathbb{E}[X].
\end{equation}

In RL, variance usually serves as a constraint, i.e., maximize $\mathbb{E}[G_0]$ and ensure $\mathbb{V}[G_0]$ is no more than a threshold~\citep{tamar2012policy,la2013actor,xie2018block}. Here, we consider the unconstrained problem by using the Lagrangian relaxation procedure~\citep{bertsekas1997nonlinear}
\begin{equation}
\label{eq:mv-obj}
    \max_\pi \mathbb{E}[G_0] - \lambda \mathbb{V}[G_0],
\end{equation}
where $\lambda$ is a trade-off hyper-parameter. Note that the mean-variance objective is in general NP-hard to optimize~\citep{mannor2011mean}. The main reason is that although the variance satisfies a Bellman equation under policy evaluation, it lacks the monotonicity of dynamic programming~\citep{sobel1982variance}. 

The policy gradient of Eq.~\ref{eq:mv-obj} is easy to get. We will refer to $\pi$, $\pi_\theta$, and $\theta$ interchangeably throughout the paper when the context is clear. Following Eq.~\ref{eq:variance_grad}, the gradient of Eq.~\ref{eq:mv-obj} is
\begin{equation}
\label{eq:mv-pg}
    \nabla_\theta \mathbb{E}[G_0]-\lambda(\nabla_\theta\mathbb{E}[G_0^2]-2\mathbb{E}[G_0]\nabla_\theta\mathbb{E}[G_0]). 
\end{equation}
Based on the analysis in Sec.~\ref{sec:pg_background}, we know that $\nabla_\theta\mathbb{E}[G_0]=\mathbb{E}[R_\tau \omega_\tau(\theta)]$, and similarly $\nabla_\theta\mathbb{E}[G_0^2]=\mathbb{E}[R_\tau^2 \omega_\tau(\theta)]$, where $R_\tau$ is the return of trajectory $\tau$ and $\omega_\tau(\theta)=\sum_t \nabla_\theta \log \pi_\theta(a_t|s_t)$.

\subsubsection{Double Sampling in Variance Policy Gradient}
In Eq.~\ref{eq:mv-pg}, the unbiased estimates for $\nabla_\theta \mathbb{E}[G_0]$ and $\nabla_\theta \mathbb{E}[G_0^2]$ can be estimated by approximating the expectations over trajectories using a single set of trajectories. In contrast, computing an unbiased estimate for $\mathbb{E}[G_0] \nabla_\theta\mathbb{E}[G_0]$ requires two distinct sets of trajectories to estimate $\mathbb{E}[G_0]$ and $\nabla_\theta \mathbb{E}[G_0]$ separately, which is known as double sampling. Some work claims that double sampling cannot be implemented without having access to a generative model of the environment that allows users to sample at least two next states~\citep{xie2018block}. This is, however, not an issue in our setting where we allow sampling multiple trajectories. As long as we get enough trajectories, estimating $\mathbb{E}[G_0] \nabla_\theta\mathbb{E}[G_0]$ is possible.

Still, different methods were proposed to avoid this double sampling. For example, \citet{tamar2012policy} produces faster estimates for $\mathbb{E}[G_0]$ and $\mathbb{V}[G_0]$ and a slower updating for $\theta$ at each episode. \citet{la2013actor} uses a perturbation method and a smoothed function method to compute the gradient of value functions with respect to $\theta$. \citet{xie2018block} introduces Fenchel duality to avoid the term $\mathbb{E}[G_0] \nabla_\theta\mathbb{E}[G_0]$ in the gradient.

\subsubsection{The Quadratic Term in Variance Policy Gradient} 
The presence of the quadratic term $R_\tau^2$ in the variance policy gradient, i.e., $\mathbb{E}_\tau[R^2_\tau\omega_\tau(\theta)]$, makes the gradient sensitive to the numerical scale of the return, and makes gradient updates unstable (later shown in experiments). This issue is inherent in all methods that require computing $\nabla_\theta\mathbb{E}[G_0^2]$ due to the use of variance. Users can not simply scale the reward by a small factor to reduce the magnitude of $R_\tau^2$, since variance is not positive homogeneous. When scaling reward by a factor $c$, $\mathbb{E}[G_0]$ is scaled by $c$ but $\mathbb{V}[G_0]$ is scaled by $c^2$. Consequently, it may lead to different optimal policies being obtained. 

In the following subsections, we discuss the remaining measures of variability mentioned in Sec.~\ref{sec:measure_var} to replace the variance in the objective of Eq.~\ref{eq:mv-obj}. We will mainly discuss the gradient derivation of those metrics.

\subsection{Gini Deviation}
Gini Deviation (GD)~\citep{gini1912variabilita}, also known as Gini mean difference or mean absolute difference, is briefly discussed in Sec.~\ref{sec:measure_var}. We repeat the definition here for clarity. For a random variable $X$, let $X^*$  be an i.i.d.~copy of $X$. Then GD is given by 
\vspace{-2pt}
\begin{equation}
\label{eq:gini}
    \mathrm{GD}[X] = \frac{1}{2}\mathbb{E}[|X - X^{*}|].
\end{equation}
\vspace{-2pt}
Variance can be defined in a similar way as 
\vspace{-2pt}
\begin{equation}
\label{eq:variance}
    \mathbb{V}[X] = \frac{1}{2}\mathbb{E}[(X - X^{*})^2]. 
\end{equation}
\vspace{-2pt}
Comparing Eq.~\ref{eq:gini} and Eq.~\ref{eq:variance}, the only difference between GD and variance is that GD uses absolute function as the difference measure (L1 distance) while variance uses quadratic function (squared L2 distance). This similarity between GD and variance makes them share some other  properties~\citep{yitzhaki2003gini}. For example, they can be presented as a weighted sum of order statistics. We refer the reader to \cite{yitzhaki2003gini} for a full overview. In addition, \cite{yitzhaki2003gini} argues that GD is superior to variance as a measure of variability for distributions far from Gaussian. There also exists numerical inequality between GD and variance, i.e.,

\begin{itemize}
\item $\mathbb{V}[X]\geq \sqrt{3}~\mathrm{GD}[X]$ for all $X\in \mathcal{M}^2$,
\end{itemize}
which is known as Glasser's inequality~\citep{glasser1962variance}.

\subsubsection{Gini Deviation Gradient Formula}
It can be intractable to compute the gradient w.r.t. the parameters of a random variable's density function through its GD by using its original definition (Eq.~\ref{eq:gini}). However, it is tractable via its quantile representation in the form of a signed Choquet integral. GD belongs to the class of signed Choquet integral with the distortion function given by Lemma~\ref{le:gini_h}.
\begin{lemma}[\cite{wang2020characterization}]
\label{le:gini_h} 
Gini deviation is a signed Choquet integral with  $h$ given by $h(\alpha)=-\alpha^2+\alpha, \alpha\in[0,1]$.
\end{lemma}
Combining Lemma~\ref{le:quantile_rep} and~\ref{le:gini_h}, $\mathrm{GD}[X]$ can be computed alternatively as
\begin{equation}
\label{eq:gini_quantile}
    \mathrm{GD}[X] = \int^1_0F^{-1}_X(1-\alpha)dh(\alpha)=\int^1_0 F^{-1}_X(\alpha)(2\alpha-1) d\alpha.
\end{equation}

\begin{proposition}
    Let Assumptions~\ref{assump:1}, \ref{assump:2}, \ref{assump:3} hold. Then
    
    \begin{equation}
    \label{eq:gini_max}
         \nabla_\theta \mathrm{GD}[X] = -\mathbb{E}_{x\sim X}\Big[\nabla_\theta\log f_X(x;\theta) \big(b+x - 2\mathbb{E}[\max\{X,x\}]\big)\Big].
    \end{equation}
\end{proposition}

\noindent\textit{Proof.} By Eq.~\ref{eq:gini_quantile}, the gradient of $\mathrm{GD}[X]$ is 
\begin{equation}
\label{eq:integral_pg}
\nabla_\theta\mathrm{GD}[X] = \int^1_0(2\alpha-1)\nabla_\theta F^{-1}_{X}(\alpha)d\alpha= \int^1_0(2\alpha-1)\nabla_\theta q_\alpha(X;\theta) d\alpha.
\end{equation}
This requires to calculate the gradient for any $\alpha$-level quantile of $X_\theta$, i.e., $\nabla_\theta q_\alpha(X;\theta)$. Based on the assumptions and the definition of the $\alpha$-level quantile, we have $\int_{-b}^{q_\alpha(X;\theta)}f_X(x;\theta)dx=\alpha$. Taking a derivative and using the Leibniz rule we obtain
\begin{equation}
\label{eq:quantile_grad_0}
    0 = \nabla_\theta \int_{-b}^{q_\alpha(X;\theta)} f_X(x;\theta) dx =\int_{-b}^{q_\alpha(X;\theta)} \nabla_\theta f_X(x;\theta) dx + \nabla_\theta q_\alpha(X;\theta)f_X\big(q_\alpha(X;\theta);\theta\big).
\end{equation}
Rearranging the term, we get
\begin{equation}
\label{eq:quantile_grad}
    \nabla_\theta q_\alpha(X;\theta)=-\int_{-b}^{q_\alpha(X;\theta)} \nabla_\theta f_X(x;\theta) dx \cdot \big[f_X\big(q_\alpha(X;\theta);\theta\big) \big]^{-1}.
\end{equation} 
Plugging back to Equation~\ref{eq:integral_pg} gives us an intermediate version of $\nabla_\theta \mathrm{GD}[X]$,
\begin{equation}
\label{eq:intermediate_pg}
    \nabla_\theta \mathrm{GD}[X] = -\int_0^1 (2\alpha-1)\int_{-b}^{q_\alpha(X;\theta)} \nabla_\theta f_X(x;\theta) dx \cdot \big[f_X\big(q_\alpha(X;\theta);\theta\big) \big]^{-1} d\alpha.
\end{equation}
To make the integral over $\alpha$ clearer, we rewrite $q_\alpha(X;\theta)$ as $F^{-1}_{X}(\alpha)$.
\begin{equation*}
    \nabla_\theta \mathrm{GD}[X] = -\int_0^1 (2\alpha-1)\int_{-b}^{F^{-1}_{X}(\alpha)} \nabla_\theta f_X(x;\theta) dx \frac{1}{f_X(F^{-1}_{X}(\alpha);\theta)} d\alpha.
\end{equation*}
Switching the integral order, we get
\begin{equation}
\begin{aligned}
    \nabla_\theta \mathrm{GD}[X]&=-\int_{-b}^b\int_{F_{X}(x)}^1 (2\alpha-1) \nabla_\theta f_X(x;\theta)\frac{1}{f_X(F^{-1}_{X}(\alpha);\theta)} d\alpha d x\\
    &=-\int_{-b}^b \nabla_\theta f_X(x;\theta) \int_{F_{X}(x)}^1 (2\alpha-1)\frac{1}{f_X(F^{-1}_{X}(\alpha);\theta)} d\alpha dx.
\end{aligned}
\end{equation}
Denote $t= F^{-1}_{X}(\alpha)$, then $\alpha = F_{X}(t)$. Here, we further change the inner integral from $d\alpha$ to $d F_{X}(t)$, i.e., $d\alpha=d F_{X}(t)=f_X(t;\theta) dt$. The integral range for $t$ is now from $F^{-1}_{X}(F_{X}(x)) = x$ to $F_{X}^{-1}(1)=b$.
\begin{equation}
\label{eq:change_variable}
\begin{aligned}
    \nabla_\theta \mathrm{GD}[X] &= -\int_{-b}^b \nabla_\theta f_X(x;\theta) \int_x^b \big(2 F_X(t)-1\big) \frac{1}{f_X(t;\theta)} d F_{X}(t) ~dx\\
      &=-\int_{-b}^b \nabla_\theta f_X(x;\theta) \int_x^b \big(2 F_{X}(t)-1\big)dt ~dx .
\end{aligned}
\end{equation}
Applying $\nabla_\theta \log(x)=\frac{1}{x}\nabla_\theta x$ to $\nabla_\theta f_X(x;\theta)$, we have
\begin{equation}
\label{eq:gd_mid_int}
\begin{aligned}
    \nabla_\theta \mathrm{GD}[X] &=-\int_{-b}^b f_X(x;\theta) \nabla_\theta \log f_X(x;\theta) \int_x^b \big(2 F_{X}(t)-1\big)dt ~dx\\
      &= - \mathbb{E}_{x\sim X} \Big[ \nabla_\theta \log f_X(x;\theta) \int_x^b \big(2F_{X}(t)-1\big)dt\Big] .
\end{aligned}
\end{equation} 
Note that $\int_x^b F_{X}(t)$ can be represented by a formula with an expectation of a max function, i.e., using integration by parts 
\begin{equation}
\label{eq:by_parts}
\begin{aligned}
    \int^b_x F_{X}(t) dt &= tF_{X}(t)|_x^b - \int_x^b t ~d F_{X}(t)\\
    &=tF_{X}(t)|_x^b - \int_x^b t f_X(t;\theta) dt\\
    &=b - x F_{X}(x) - \int_x^b t f_X(t;\theta) dt \\
    &=b - x \int_{-b}^x f_X(t;\theta) dt - \int^b_x t f_X(t;\theta) dt \\
    &= b - \int_{-b}^b \max\{t,x\} f_X(t;\theta) dt \\
    &=b - \mathbb{E}[\max\{X, x\}] .
\end{aligned}
\end{equation}
Plugging Eq.~\ref{eq:by_parts} back to Eq.~\ref{eq:gd_mid_int}, we get Eq.~\ref{eq:gini_max}.  $\square$

\textit{Unbiased Estimator.} Next, we consider the unbiased estimator of Eq.~\ref{eq:gini_max}. Eq.~\ref{eq:gini_max} can be split into two parts as
\begin{equation*}
\nabla_\theta \mathrm{GD}[X] = -\mathbb{E}_{x\sim X}\Big[\nabla_\theta \log f_X(x;\theta)(b+x)\Big] + 2\mathbb{E}_{x\sim X}\Big[\nabla_\theta \log f_X(x;\theta) \mathbb{E}[\max\{X, x\}]\Big] .
\end{equation*}
The second part can be interpreted as
\vspace{-2pt}
\begin{equation}
\label{eq:second_part}
2\mathbb{E}\Big[\nabla_\theta\log f_X(X;\theta) \mathbb{E}[\max\{X^*, X\}]\Big] , 
\end{equation}
\vspace{-3pt}
where $X^*$ and $X$ are independent and $X^*\overset{d}{=}X$. An intuitive way to compute an unbiased estimator for Eq.~\ref{eq:second_part} is to use two distinct sample sets to estimate the two expectation terms separately. However, it is also possible if we wish to compute using a single sample set. Here we can treat one sample as coming from $X$, and the remaining samples as coming from $X^*$. Thus, given $n$ samples $\{x_i\}_{i=1}^n$ of $X$, the unbiased estimator for Eq.~\ref{eq:second_part} is
\begin{equation}
    \frac{2}{n} \sum_{i=1}^n \nabla_\theta \log f_X(x_i;\theta) \frac{1}{n-1}\sum_{j\neq i} \max\{x_j, x_i\}.
\end{equation}
Combining with the first part, the unbiased estimator for $\nabla_\theta \mathrm{GD}[X]$ is
\begin{equation}
\label{eq:gini_unbiased}
\begin{aligned}
    \nabla_\theta\mathrm{GD}[X]_{[n]} &= -\frac{1}{n} \sum_{i=1}^n \nabla_\theta\log f_X(x_i;\theta) (b+x_i) + \frac{2}{n} \sum_{i=1}^n \nabla_\theta \log f_X(x_i) \frac{1}{n-1}\sum_{j\neq i} \max\{x_j, x_i\}\\
    &= \frac{1}{n} \sum_{i=1}^n \nabla_\theta\log f_X(x_i;\theta)  \big[ \frac{2}{n-1}\sum_{j\neq i} \max\{x_j, x_i\} -(b+x_i) \big].
\end{aligned}
\end{equation}

\subsubsection{Gini Deviation Policy Gradient via Sampling}

In a typical application, $X$ in Eq.~\ref{eq:gini_max} would correspond to the performance of a system, e.g., the total return $G_0$ in RL. Note that in order to compute Eq.~\ref{eq:gini_unbiased}, one needs access to $\nabla_\theta \log f_X(x;\theta)$: the sensitivity of the system performance to the parameters $\theta$. Usually, the system performance is a complicated function, and calculating its probability distribution is intractable. However, in RL, the performance is a function of trajectories. The sensitivity of the trajectory distribution is often easy to compute. This naturally suggests a sampling based algorithm for policy gradient estimation.

Now consider the unbiased estimator (Eq.~\ref{eq:gini_unbiased}) in the context of RL, i.e., $X=G_0$ and $\theta$ is the policy parameter. We have
\vspace{-2pt}
\begin{equation}
    \nabla_\theta \mathrm{GD}[G_0] = \frac{1}{n} \sum_{i=1}^n \nabla_\theta \log f_{G_0}(g_i;\theta) \big[  \frac{2}{n-1}\sum_{j\neq i} \max\{g_j,g_i\} -(b+g_i)\big],
\end{equation}
\vspace{-3pt}
where $\{g_i\}_{i=1}^n$ are samples from $G_0$. In practice, to sample from $G_0$, we need to sample a trajectory $\tau=(s_0,a_0,r_1,s_1,a_1,..., r_T,s_T)$ from the environment by executing $\pi_\theta$ and then compute its corresponding return $R_\tau := r_1 + \gamma r_2 + ... + \gamma^{T-1}r_{T}$. The probability of the sampled return can be calculated as $f_{G_0}(R_\tau;\theta) =\mu_0(s_0)\prod_{t=0}^{T-1}[\pi_\theta(a_t|s_t)p( r_{t+1}|s_t,a_t)]$. The gradient of its log-likelihood is the same as that of $P(\tau|\theta)=\mu_0(s_0)\prod_{t=0}^{T-1}[\pi_\theta(a_t|s_t)p( s_{t+1}|s_t,a_t)]$, since the difference in transition probability does not alter the policy gradient. According to Sec.~\ref{sec:pg_background}, $\nabla_\theta P(\tau|\theta)=\sum_{t=0}^{T-1}\nabla_\theta \log \pi_\theta(a_{t}|s_{t})$.

Thus, given trajectory samples $\{\tau_i\}_{i=1}^n$ (with corresponding trajectory returns $\{R_{\tau_i}\}_{i=1}^n$), an unbiased estimation for GD policy gradient is
\begin{equation}
\label{eq:gini_est}
\begin{aligned}
    \nabla_\theta\mathrm{GD}[G_0] &= \frac{1}{n} \sum_{i=1}^n \eta_i \sum_{t=0}^{T-1}\nabla_\theta\log \pi_\theta(a_{i,t}|s_{i,t}), \\
    \mathrm{where} ~\eta_i&=\frac{2}{n-1}\sum_{i\neq j}\max\{R_{\tau_j},R_{\tau_i}\} - (b+R_{\tau_i}) .
\end{aligned}
\end{equation}

Note that the policy gradient calculation bears similarity in the remaining metrics. We may give the formulas without detailed explanations. 

\subsection{Mean Deviation}
Mean Deviation (MD) is defined as
\begin{equation}
    \mathrm{MD}[X] = \mathbb{E}[|X - \mathbb{E}[X]|],
\end{equation}
which describes the average L1 distance from the mean.
\subsubsection{Mean Deviation Gradient Formula}
\begin{restatable}{proposition}{mdgrad}
    Let Assumptions~\ref{assump:1}, \ref{assump:2}, \ref{assump:3} hold. Then
    \begin{equation}
        \label{eq:md_grad}\nabla_\theta\mathrm{MD}[X]=\mathbb{E}_{x\sim X}\Big[|x-\mathbb{E}[X]| \nabla_\theta \log f_X(x;\theta)\Big] - \mathbb{E}_{x\sim X}\Big[\mathrm{sgn}(x-\mathbb{E}[X])\Big]\nabla_\theta\mathbb{E}[X] ,
    \end{equation}
\end{restatable}
\noindent Proof see Appendix~\ref{sec:mdgrad-proof}. Here, we remain $\nabla_\theta\mathbb{E}[X]$ in the equation since it is easy to estimate in the RL setting.

\textit{Unbiased Estimator.} Similar to the case in variance, double sampling is required to estimate $\nabla_\theta\mathbb{E}[X]$ in the second term. For simplicity, we keep the $\nabla_\theta \mathbb{E}[X]$ in the equation and consider how to construct an unbiased estimator for the remaining terms using a single set of samples. Analogous to GD (Eq.~\ref{eq:second_part}), we may interpret the remaining terms as 
\begin{equation}
    \mathbb{E}\Big[|X-\mathbb{E}[X^*]|\nabla_\theta \log f_X(X;\theta)\Big] ~~\mathrm{and}~~ \mathbb{E}\Big[\mathrm{sgn}(X-\mathbb{E}[X^*])\Big] ,
\end{equation}
where $X^*$ and $X$ are independent  and $X^*\overset{d}{=}X$. Given $n$ samples $\{x_i\}^n_{i=1}$ of $X$, the unbiased estimator for them are
\begin{equation}
\label{eq:md-terms}
    \frac{1}{n}\sum_{i=1}^n \Big|x_i - \frac{1}{n-1}\sum_{j\neq i}x_j\Big| \nabla_\theta \log f_X(x_i;\theta)~~\mathrm{and}~~ \frac{1}{n} \sum_{i=1}^n \mathrm{sgn}(x_i - \frac{1}{n-1}\sum_{j\neq i} x_j) .
\end{equation}

\subsubsection{Mean Deviation Policy Gradient via Sampling}
In the context of RL, e.g., computing $\nabla_\theta \mathrm{MD}[G_0]$, given trajectory samples $\{\tau_i\}_{i=1}^n$ (with corresponding trajectory returns $\{R_{\tau_i}\}_{i=1}^n$), the terms in Eq.~\ref{eq:md-terms} are computed as 
\begin{equation}
    \frac{1}{n}\sum_{i=1}^n|R_{\tau_i}-\frac{1}{n-1}\sum_{j\neq i}R_{\tau_j}| \sum_{t=0}^{T-1}\nabla_\theta\log\pi_\theta(a_{i,t}|s_{i,t})~\mathrm{and}~\frac{1}{n}\sum_{i=1}^n\mathrm{sgn}(R_{\tau_i}-\frac{1}{n-1}\sum_{j\neq i}R_{\tau_j}) .
\end{equation}
Therefore, the policy gradient $\nabla_\theta\mathrm{MD}[G_0]$ is estimated as
\begin{equation}
\begin{aligned}
\nabla_\theta \mathrm{MD}[G_0]&=\frac{1}{n}\sum_{i=1}^n|\eta_i| \sum_{t=0}^{T-1}\nabla_\theta\log\pi_\theta(a_{i,t}|s_{i,t})-\frac{1}{n}\sum_{i=1}^n\mathrm{sgn}(\eta_i) \nabla_\theta\mathbb{E}[G_0], \\
\mathrm{where}~\eta_i &= R_{\tau_i}-\frac{1}{n-1}\sum_{j\neq i}R_{\tau_j}. 
\end{aligned}
\end{equation}
We need another set of trajectory samples to estimate $\nabla_\theta\mathbb{E}[G_0]$.

\subsection{Mean-Median Deviation}
Mean-Median Deviation (MMD) is defined as 
\begin{equation}
\label{eq:mmd}
   \mathrm{MMD}[X]=\mathbb{E}[|X-\mathrm{Median(X)}|]=\min_{x\in\mathbb{R}}\mathbb{E}[|X-x|], 
\end{equation}
which describes the average L1 distance from the median.

\subsubsection{Mean-Median Deviation Gradient Formula}
The appearance of the minimum in Eq.~\ref{eq:mmd} can complicate the optimization. Here, we switch to its quantile representation in the form of signed Choquet integral. MMD belongs to the class of signed Choquet integral with the distortion function given by Lemma~\ref{le:mmd_h}.

\begin{lemma}[\cite{wang2020distortion}]
\label{le:mmd_h} 
The Mean-Median Deviation is a signed Choquet integral with $h$ given by $h(\alpha) = \min\{\alpha, 1-\alpha\}$.
\end{lemma}
Combining Lemma~\ref{le:quantile_rep} and~\ref{le:mmd_h}, $\mathrm{MMD}[X]$ can be computed alternatively as
\begin{equation}
\begin{aligned}
    \mathrm{MMD}[X] &= \int_0^1 F_X^{-1}(1-\alpha) d \min\{\alpha, 1-\alpha\}\\
    &=\int_0^{\frac{1}{2}} F^{-1}_X(1-\alpha) d\alpha + \int^1_{\frac{1}{2}} F^{-1}_X(1-\alpha) d(1-\alpha)\\
    &= \int^1_{\frac{1}{2}} F^{-1}_X(\alpha) d\alpha - \int_0^{\frac{1}{2}} F^{-1}_X(\alpha) d\alpha .
\end{aligned}  
\end{equation}

\begin{restatable}{proposition}{mmdgrad}
    Let Assumptions~\ref{assump:1}, \ref{assump:2}, \ref{assump:3} hold. Then
    \begin{equation}
        \label{eq:mmd_grad}
        \begin{aligned}\nabla_\theta\mathrm{MMD}[X]&=\frac12 \mathbb{E}\Big[\nabla_\theta\log f_X(X;\theta)(2q_{\frac12}(X;\theta)-X-b)| X\leq q_{\frac12}(X;\theta)\Big]\\
    &-\frac12 \mathbb{E}\Big[\nabla_\theta \log f_X(X;\theta)(b-X)|X\geq q_{\frac12}(X;\theta)\Big] , 
    \end{aligned}
    \end{equation}
    where $q_{\frac{1}{2}}(X;\theta)$ is the half quantile of $X$.
\end{restatable}
\noindent Proof see Appendix~\ref{sec:mmdgrad-proof}.

\textit{Biased Estimator.} Without access to the true underlying distribution of $X$, we have to estimate an empirical half quantile from samples. Given n samples $\{x_i\}_{i=1}^n$ of $X$, the empirical $\frac{1}{2}$-quantile is
\begin{equation}
    \hat{q}_{\frac12} = \inf_z \hat{F}(z)\geq \frac12 ,
\end{equation}
where $\hat{F}$ is the empirical cdf of $X$: $\hat{F}(z)=\frac{1}{n}\sum_{i=1}^n\mathbb{I}_{x_i}\leq z$. Therefore, a practical estimator of the MMD gradient is
\begin{equation}
\label{eq:mmd_grad_est}
\begin{aligned}
\nabla_\theta\widehat{\mathrm{MMD}}[X]_{[n]} &\simeq \frac12 \frac{1}{n/2}\sum_{i=1}^n(2\hat{q}_{\frac12}-x_i-b)\nabla_\theta \log f_X(x_i;\theta)\mathbb{I}_{x_i\leq \hat{q}_{\frac12}}\\
&+\frac12 \frac{1}{n/2}\sum_{i=1}^n(b-x_i)\nabla_\theta\log f_X(x_i;\theta)\mathbb{I}_{x_i\geq \hat{q}_{\frac12}} .
\end{aligned}
\end{equation}

Since the empirical half quantile is a biased estimator of $q_{\frac12}(X;\theta)$,  $\nabla_\theta\widehat{\mathrm{MMD}}_{[n]}[X]$ is also a biased estimator. However, it is a consistent estimator and the bias can be bounded. The proof is similar to the proof of Theorem 4.1 in \cite{hong2009simulating} and the proof of Theorem 3 in \cite{tamar2015optimizing}.

\begin{restatable}{theorem}{mmdconsist} \label{thm:mmd-consis}
    Let Assumptions~\ref{assump:1}, \ref{assump:2}, \ref{assump:3} hold. Then $\nabla_{\theta}\widehat{\mathrm{MMD}}[X]_{[n]}\rightarrow \nabla_\theta \mathrm{MMD}[X]$ as $n\rightarrow \infty$.
\end{restatable}
\noindent Proof see Appendix~\ref{sec:thm1-proof}.

The following assumption is required to prove the error bound by following \cite{hong2009simulating}. Denote $D_1(X;\theta)=(2q_{\frac12}(X;\theta)-X-b)\nabla_\theta \log f_X(X;\theta)$, and $g_1(y;\theta)=\mathbb{E}[D_1(X;\theta)|X=y]$. Denote $D_2(X;\theta)=(b-X)\nabla_\theta \log f_X(X;\theta)$, and $g_2(y;\theta)= \mathbb{E}[D_2(X;\theta)|X=y]$.
\begin{assumption}
\label{assump:4}
$g_1(y;\theta)$ and $g_2(y;\theta)$ are continuous at $q_{\frac12}(X;\theta)$ and $f_X(q_{\frac12}(X;\theta);\theta)>0$, for all $\theta$.
\end{assumption}
\begin{restatable}{theorem}{mmdbound}
\label{thm:mmd-bound}
Let Assumptions~\ref{assump:1}, \ref{assump:2}, \ref{assump:3} ,\ref{assump:4} hold. Then $\mathbb{E}\big[\nabla_{\theta}\widehat{\mathrm{MMD}}[X]_{[n]}\big] - \nabla_\theta \mathrm{MMD}[X]$ is $O(n^{-\frac{1}{2}})$.
\end{restatable}
\noindent Proof see Appendix~\ref{sec:thm2-proof}.
\subsubsection{Mean-Median Deviation Policy Gradient via Sampling}

In the context of RL, e.g., computing $\nabla_\theta \mathrm{MMD}[G_0]$, given trajectory samples $\{\tau_i\}_{i=1}^n$(with corresponding trajectory returns $\{R_{\tau_i}\}_{i=1}^n$), the gradient is computed as
\begin{equation}
\begin{aligned}
    \nabla_\theta \mathrm{MMD}[G_0] \simeq & ~~\frac12 \frac{1}{n/2}\sum_{i=1}^n(2 \hat{q}_{\frac{1}{2}}- R_{\tau_i} -b)\mathbb{I}_{R_{\tau_i}\leq \hat{q}_{\frac{1}{2}}} \sum_{t=0}^{T-1} \nabla_\theta \log\pi_\theta(a_{i,t}|s_{i,t}) \\
    & - \frac12 \frac{1}{n/2}\sum_{i=1}^n (b-R_{\tau_i})\mathbb{I}_{R_{\tau_i}\geq \hat{q}_{\frac12}}\sum_{t=0}^{T-1}\nabla_\theta \log \pi_\theta(a_{i,t}|s_{i,t}) , \\
\end{aligned}
\end{equation}
where $\hat{q}_{\frac12}$ is the empirical $\frac{1}{2}$-quantile estimated from $\{R_{\tau_i}\}_{i=1}^n$.

\subsection{Standard Deviation}

Standard Deviation (STD) is defined as

\begin{equation}
    \mathrm{STD}[X] = \sqrt{\mathbb{V}[X]} = \sqrt{\mathbb{E}[X^2]-(\mathbb{E}[X])^2},
\end{equation}
STD can also be defined similarly to signed Choquet integral as
\begin{equation}
    \mathrm{STD}[X]=\sup_{h\in\mathcal{H}'} \int_0^1 F^{-1}_X(1-\alpha) dh(\alpha) , 
\end{equation}
where $\mathcal{H}'=\{h\in\mathcal{H}, h(1)=0,\int_0^1(h'(t))^2 dt<1,h~\mathrm{is~concave}\}$~\citep{wang2020characterization}. But this formulation may complicate the optimization due to the supreme over a function space. The gradient of STD is easy to compute through the gradient of variance. Thus, we keep this subsection short as the gradient of variance is discussed in Sec.~\ref{sec:mean-variance-rl}.

\begin{equation}
    \nabla_\theta\mathrm{STD}[X]=\frac{1}{2\sqrt{\mathbb{V}[X]}}\nabla_\theta \mathbb{V}[X] .
\end{equation}

The policy gradient of $\mathrm{STD}[G_0]$ follows the policy gradient of $\mathbb{V}[G_0]$ by dividing it by $2\sqrt{\mathbb{V}[G_0]}$.


\subsection{Inter-Quantile Range}
Inter-Quantile Range (IQR) is defined as 
\begin{equation}
    \label{eq:iqr_def}\mathrm{IQR}_\alpha[X]=F^{-1}_X(\alpha) - F^{-1}_X(1-\alpha), \alpha\in[\frac{1}{2},1),
\end{equation}
which is the difference between two quantile values at quantile levels $\alpha$ and $1-\alpha$ (the levels are symmetric w.r.t $0.5$). When $\alpha \rightarrow 1$, IQR recovers the full range of $X$.

\subsubsection{Inter-Quantile Range Gradient Formula}
IQR also belongs to the class of signed Choquet integral
with the distortion function given by $h(t) = \mathbb{I}_{\{1-\alpha\leq t \leq \alpha\}}$~\citep{wang2020distortion}. However, IQR can be directly optimized through its original definition. 

\begin{restatable}{proposition}{iqrgrad}
    Let Assumptions~\ref{assump:1}, \ref{assump:2}, \ref{assump:3} hold. Then
    \begin{equation}
    \begin{aligned}
        \nabla_\theta \mathrm{IQR}_\alpha[X]=&-\frac{\alpha}{f_X(q_\alpha(X;\theta);\theta)} \mathbb{E}\Big[\nabla_\theta \log f_X(x;\theta)|X\leq q_\alpha(X;\theta)\Big] \\
    &+\frac{1-\alpha}{f_X(q_{1-\alpha}(X;\theta);\theta)} \mathbb{E}\Big[\nabla_\theta \log f_X(x;\theta)|X\leq q_{1-\alpha}(X;\theta)\Big] ,
    \end{aligned}
    \end{equation}
\end{restatable}
\noindent Proof see Appendix~\ref{sec:iqrgrad-proof}.

\textit{Biased Estimator.} Similar to MMD, we have to estimate the empirical $\alpha$ and $1-\alpha$ quantile from samples, i.e., $\hat{q}_\alpha$ and $\hat{q}_{1-\alpha}$. In addition, the access to the density function $f_X$ is required to compute the denominator. This is, however, generally unavailable since we made no assumption on the return distribution class. The estimation of this density function is discussed by several quantile-based policy gradient papers. For example, \cite{jiang2023quantile} omitted the denominator $f_X$ as they argued that the remaining gradient of quantile still follows the same direction without this factor. However, this does not apply to the gradient of IQR since its gradient consists of two quantile gradients, omitting the denominator alters the gradient update direction. \cite{jiang2022quantile} proposed two approaches to estimate $f_X$. One is the well-known kernel density estimation (KDE) \citep{scott2015multivariate}. The other is to approximate $f_X$ by a sigmoid function. We choose the KDE approach in our experiments. Specifically, we use \texttt{scipy.stats.gaussian\_kde} with the ``silverman" bandwidth method in our implementation. Note that although the estimation bias of empirical quantile can be bounded, the accuracy of the density estimation highly depends on the sample size and the chosen density estimation algorithm.

Given $n$ samples $\{x_i\}_{i=1}^n$ of $X$, a practical estimator for IQR gradient is

\begin{equation}
\begin{aligned}
    \nabla_\theta \mathrm{IQR}_\alpha[X]_{[n]}\simeq &-\frac{\alpha}{KDE(\hat{q}_\alpha)} \frac{1}{\alpha n} \sum_{i=1}^n \nabla_\theta \log f_X(x_i;\theta) \mathbb{I}_{x_i\leq \hat{q}_\alpha}\\
    &+\frac{1-\alpha}{KDE(\hat{q}_{1-\alpha})}\frac{1}{(1-\alpha)n}\sum_{i=1}^n\nabla_\theta\log f_X(x_i;\theta) \mathbb{I}_{x_i\leq \hat{q}_{1-\alpha}} .
\end{aligned}
\end{equation}
\subsubsection{Inter-Quantile Range Policy Gradient via Sampling}
In the context of RL, e.g., comparing $\nabla_\theta\mathrm{IQR}[G_0]$, given trajectory samples $\{\tau\}_{i=1}^n$ (with corresponding trajectory returns $\{R_{\tau_i}\}_{i=1}^n$), the gradient is computed as
\begin{equation}
\begin{aligned}
    \nabla_\theta \mathrm{IQR}_\alpha[G_0]\simeq &-\frac{1}{KDE(\hat{q}_\alpha)}\frac{1}{n} \sum_{i=1}^n \mathbb{I}_{R_{\tau_i}\leq \hat{q}_\alpha} \sum_{t=0}^{T-1}\nabla_\theta \log \pi_\theta(a_{i,t}|s_{i,t})\\
    &+ \frac{1}{KDE(\hat{q}_{1-\alpha})}\frac{1}{n} \sum_{i=1}^n \mathbb{I}_{R_{\tau_i}\leq \hat{q}_{1-\alpha}}\sum_{t=0}^{T-1}\pi_\theta(a_{i,t}|s_{i,t}) ,
\end{aligned}
\end{equation}
where $\hat{q}_{\alpha}$ and $\hat{q}_{1-\alpha}$ are the empirical $\alpha$ and $1-\alpha$ level quantile estimated from $\{R^{\tau_i}\}_{i=1}^n$.

\subsection{CVaR Deviation}
\label{sec:cvar-dev}

The lower tail CVaR Deviation is
\begin{equation}
\mathrm{CD}[X]=\mathbb{E}[X] -\mathrm{CVaR}^\lor_\alpha(X),
\end{equation}
which describes the deviation of the left tail value from the mean.

\subsubsection{CVaR Deviation Gradient Formula}
\begin{restatable}{proposition}{cdgrad}
    Let Assumptions~\ref{assump:1}, \ref{assump:2}, \ref{assump:3} hold. Then
    \begin{equation}
        \nabla_\theta\mathrm{CD}[X]=\nabla_\theta \mathbb{E}[X] - \mathbb{E}[(X-q_\alpha(X;\theta))\nabla_\theta \log f_X(X;\theta)|X\leq q_\alpha(X;\theta)] .
    \end{equation}
\end{restatable}
\noindent Proof see Appendix~\ref{sec:cdgrad-proof}. Here we remain $\nabla_\theta\mathbb{E}[X]$ in the equation since it is easy to estimate in RL setting.

\textit{Biased Estimator.} For the CVaR gradient, i.e., $\nabla_\theta\mathrm{CVaR}^\lor_\alpha(X)$, in CVaR Deviation, similar to MMD and IQR, we need to estimate an empirical $\alpha$-quantile from samples, i.e., $\hat{q}_\alpha$. Given $n$ samples $\{x_i\}_{i=1}^n$ of X, a practical estimator for CVaR gradient is
\begin{equation}
    \nabla_\theta \widehat{\mathrm{CVaR}}_\alpha ^\lor(X)_{[n]} = \frac{1}{\alpha n} \sum_{i=1}^n (x_i - \hat{q}_\alpha)\mathbb{I}_{x_i\leq \hat{q}_\alpha} \nabla_\theta \log f_X(x_i;\theta) .
\end{equation}
The empirical $\alpha$-quantile also introduces bias to the estimator, but it is a consistent estimator and the bias can be bounded.
\begin{theorem}[\cite{tamar2015optimizing}]
\label{thm:cvar-consis} 
\raggedright Let Assumptions~\ref{assump:1}, \ref{assump:2}, \ref{assump:3} hold. Then $\nabla_{\theta}\widehat{\mathrm{CVaR}}^\lor_\alpha(X)_{[n]}\rightarrow \nabla_\theta \mathrm{CVaR}^\lor_\alpha(X)$ as $n\rightarrow \infty$.
\end{theorem}

The following assumption is required to prove the error bound by following \cite{hong2009simulating}. Denote $D(X;\theta)=(X-q_\alpha(X;\theta))\nabla_\theta \log f_X(X;\theta)$ and $g(y;\theta)=\mathbb{E}[D(X;\theta)|X=y]$.

\begin{assumption}
\label{assump:5}
$g(y;\theta)$  is continuous at $q_{\alpha}(X;\theta)$ and $f_X(q_{\alpha}(X;\theta);\theta)>0$, for all $\theta$.
\end{assumption}

\begin{theorem}[\cite{tamar2015optimizing}]
\raggedright Let Assumptions~\ref{assump:1}, \ref{assump:2}, \ref{assump:3} ,\ref{assump:5} hold. Then $\mathbb{E}\big[\nabla_{\theta}\widehat{\mathrm{CVaR}}^\lor_\alpha(X)_{[n]}] - \nabla_\theta \mathrm{CVaR}^\lor_\alpha(X)$ is $O(n^{-\frac{1}{2}})$.
\end{theorem}

\subsubsection{CVaR Deviation Policy Gradient via Sampling}

In the context of RL, e.g., computing $\nabla_\theta \mathrm{CD}[G_0]$, given trajectory samples $\{\tau_i\}_{i=1}^n$ (with corresponding trajectory returns $\{R_{\tau_i}\}_{i=1}^n$), the gradient is computed as

\begin{equation}
    \nabla_\theta \mathrm{CD}[G_0]\simeq\frac{1}{n}\sum_{i=1}^n R_{\tau_i} \sum_{t=0}^{T-1}\nabla_\theta \log \pi_\theta(a_{i,t}|s_{i,t}) - \frac{1}{\alpha n}\sum_{i=1}^n (R_{\tau_i}-\hat{q}_\alpha)\mathbb{I}_{R_{\tau_i}\leq \hat{q}_\alpha} \sum_{t=0}^{T-1} \nabla_\theta \log \pi_\theta(a_{i,t}|s_{i,t}) ,
\end{equation}
where $\hat{q}_\alpha$ is the empirical $\alpha$-level quantile estimated from $\{R_{\tau_i}\}_{i=1}^n$.

\subsection{Semi\_Variance}
\label{sec:semi_var}
Semi\_Variance (SV) is closely related to variance but only computes the dispersion of the distribution that falls below or above the mean. Here, we choose the downside SV as an example. In this paper, we adopt SV defined as~\citep{ma2022mean}

\begin{equation}
\label{eq:sv-def}
    {\mathrm{S}}\mathbb{V}[X] = \mathbb{E}[(X-\mathbb{E}[X])^2 \mathbb{I}_{X\leq\mathbb{E}[X]}] .
\end{equation}



\subsubsection{Semi-Variance Gradient Formula}

\begin{restatable}{proposition}{svargrad}
    Let Assumptions~\ref{assump:1}, \ref{assump:2}, \ref{assump:3} hold. Then
    \begin{equation}
        \nabla_\theta \mathrm{S}\mathbb{V}[X]=\mathbb{E}\Big[(X-\mathbb{E}[X])^2 \nabla_\theta\log f_X(X;\theta)\mathbb{I}_{X\leq \mathbb{E}[X]}\Big] + \nabla_\theta \mathbb{E}[X] \cdot\mathbb{E}\Big[2(\mathbb{E}[X]-X)\mathbb{I}_{X\leq \mathbb{E}[X]}\Big].
    \end{equation} 
\end{restatable}
\noindent Proof see Appendix~\ref{sec:svargrad-proof}

\textit{Unbiased Estimator.}
Similar to the case in variance, double sampling is required to estimate $\nabla_\theta\mathbb{E}[X]$ in the second term. We remain $\nabla_\theta\mathbb{E}[X]$ in the equation and consider how to estimate the remaining terms using a single set of samples. We may interpret the remaining terms as
\begin{equation}
\label{eq:sv-item}
    \mathbb{E}\Big[(X-\mathbb{E}[X^*])^2 \nabla_\theta \log f_X(X;\theta) \mathbb{I}_{{X\leq\mathbb{E}[X^*]}}\Big]~~\mathrm{and}~~ \mathbb{E}\Big[2(\mathbb{E}[X^*]-X) \mathbb{I}_{X\leq \mathbb{E}[X^*]}\Big] , 
\end{equation}
where $X^*$ and $X$ are independent  and $X^*\overset{d}{=}X$. Given $n$ samples $\{x_i\}_{i=1}^n$ of $X$, the unbiased estimator for them are 
\begin{equation}
\label{eq:sv-terms}
    \frac{1}{n}\sum_{i=1}^n (x_i - y_i)^2 \nabla_{\theta} f_X(x_i;\theta) \mathbb{I}_{\{x_i\leq y_i\}}~~\mathrm{and}~~ \frac{1}{n}\sum_{i=1}^n 2(y_i-x_i)\mathbb{I}_{\{x_i\leq y_i\}} , 
\end{equation}
where $y_i = \frac{1}{n-1}\sum_{j\neq i}x_j$.

\subsubsection{Semi-Variance Policy Gradient via Sampling}
In the context of RL, e.g., computing $\nabla_\theta \mathrm{S}\mathbb{V}[G_0]$, given trajectory samples $\{\tau_i\}_{i=1}^n$ (with corresponding trajectory returns $\{R_{\tau_i}\}_{i=1}^n$), the gradient is computed as
\begin{equation}
\begin{aligned}
    \nabla_\theta {\mathrm{S}}\mathbb{V}[G_0] &= \frac{1}{n}\sum_{i=1}^n\eta_i^2 \mathbb{I}_{\{R_{\tau_i}\leq \bar{R}_{\tau_{-i}}\}}\sum_{t=0}^{T-1}\nabla_\theta \pi_\theta(a_{i,t}|s_{i,t}) + \frac{1}{n}\sum_{i=1}^n 2\eta_i\mathbb{I}_{\{R_{\tau_i}\leq\bar{R}_{\tau_{-i}}\}} \cdot \nabla_\theta \mathbb{E}[G_0] ,\\
    \mathrm{where} ~\eta_i&=\bar{R}_{\tau_{-i}}-R_{\tau_i}, ~\mathrm{and}~\bar{R}_{\tau_{-i}}=\frac{1}{n-1}\sum_{j\neq i} R_{\tau_j}.
\end{aligned}
\end{equation}
We need another set of trajectory samples to estimate $\nabla_\theta \mathbb{E}[G_0]$

\subsection{Semi\_Standard Deviation}
Semi\_STD or simply Semi\_Deviation (SD) is the square root of SV. Considering the downside of the distribution, SD is defined as
\begin{equation}
    \mathrm{SD}[X]=\sqrt{{\mathrm{S}}\mathbb{V}[X]} = (\mathbb{E}[(X-\mathbb{E}[X])^2 \mathbb{I}_{X\leq\mathbb{E}[X]}])^\frac12 .
\end{equation}

The gradient of SD is easy to compute through the gradient of SV. Thus we keep this subsection short. The gradient of SD is also discussed by \cite{tamar2015policy}. 
\begin{equation}
    \nabla_\theta \mathrm{SD}[X] = \frac{1}{2\sqrt{{\mathrm{S}}\mathbb{V}[X]}} \nabla_\theta {\mathrm{S}}\mathbb{V}[X] .
\end{equation}

The policy gradient of $\mathrm{SD}[G_0]$ follows the policy gradient of ${\mathrm{S}}\mathbb{V}[G_0]$ by dividing it by $2\sqrt{{\mathrm{S}}\mathbb{V}[G_0]}$.

\section{Mean-Variability Policy Gradient Algorithm}

With the metrics discussed in this paper, we consider a broader mean-variability problem, i.e.,
\begin{equation}
    \max_\pi \mathbb{E}[G_0] - \lambda \mathbb{D}[G_0] , 
\end{equation}
where $\lambda$ is a trade-off parameter; $\mathbb{D}[G_0]$ denotes the variability of the policy's return and can be instantiated to the metrics we have discussed. 

$\mathbb{E}[G_0]$ can be maximized by well-established on-policy RL algorithms, e.g., REINFORCE with baseline introduced in Sec.~\ref{sec:pg_background} or PPO~\citep{schulman2017proximal}. We describe the algorithm of the mean-variability policy gradient in Algo.~\ref{algo:rf_variability}, where the mean part is updated by REINFORCE. Note that for gradients requiring double sampling, we randomly split the sampled trajectories into two sets for gradient calculation.

\begin{algorithm}[t]
        \caption{Mean-Variability Policy Gradient (with REINFORCE baseline)}
        \SetAlgoLined
        \DontPrintSemicolon
        \KwIn{Iterations $K$, sample size $n$, policy learning rate $\alpha_\theta$, value learning rate $\alpha_\phi$, trade-off parameter $\lambda$;}
        \textbf{Initialize:}  policy $\pi_\theta$, value function $V_\phi$\\
        \For{$k$ in $1:K$}{
            \tcp{\textcolor{blue}{Sample trajectories}}
            $\{\tau_i\}_{i=1}^n \leftarrow$ run\_episodes($\pi_\theta,n$)
            
            Compute trajectory return $R_{\tau_i}$ and reward-to-go $\{R_{\tau_i,t}\}_{t=0}^{T-1}$ for each $\tau_i$
            
            \tcp{\textcolor{blue}{Mean PG, e.g., REINFORCE with baseline  } }
            
            Compute mean\_grad via Eq.~\ref{eq:rf_est}
            
            \tcp{\textcolor{blue}{Variability PG}}
            Split $\{\tau_i\}_{i=1}^n$ for double sampling if required
            
            Compute variability\_grad via corresponding equation in Sec.~\ref{sec:other-measures}
            
            \tcp{\textcolor{blue}{Update policy}}
            $\theta \leftarrow \theta + \alpha_\theta~ (\mathrm{mean\_grad} - \lambda~ \mathrm{variability\_grad})$
            
            \tcp{\textcolor{blue}{Update value function}}
            
            \For{$i$ in $1:n$}{
            $\phi\leftarrow \phi - \alpha_\phi \nabla_\phi \frac{1}{T}\sum_{t=0}^{T-1}(V_\phi(s_{i,t})-R_{\tau_i,t})^2$
            }
            
        }
\label{algo:rf_variability}
\end{algorithm}

For on-policy policy gradient, samples are abandoned once the policy is updated, which is expensive for gradient calculation since we are required to sample $n$ trajectories each time. To improve the sample efficiency to a certain degree, we may consider incorporating importance sampling (IS) to reuse samples for multiple updates in each loop. For each $\tau_i$, the IS ratio is $\rho_i=\prod_{t=0}^{T-1}\pi_\theta(a_{i,t}|s_{i,t})/\pi_{\hat{\theta}}(a_{i,t}|s_{i,t})$, where $\theta$ is the current policy parameter and $\hat{\theta}$ is the old policy parameter when $\{\tau_i\}_{i=1}^n$ are sampled.

The extreme IS values $\rho_i$ will introduce high variance to the policy gradient. To stabilize learning, there are several commonly used techniques. For example, one strategy is to ignore samples with extreme ratios to prevent them from dominating the estimate. Another common strategy is weight clipping by directly clipping the extreme ratios to a small value. In this paper, we adopt the weight clipping scheme for its simplicity. We clip $\rho_i$ by a constant value $\zeta$, i.e., $\rho_i=\min(\rho_i, \zeta)$, though it will introduce bias, e.g., see~\cite{bottou2013counterfactual}. When IS is incorporated, it is also convenient to work with PPO (where IS is already included). We describe the algorithm of the mean-variability policy gradient with IS using PPO in Algo.~\ref{algo:ppo_variability}.

\begin{algorithm}[H]
        \caption{Mean-Variability Policy Gradient (with PPO)}
        \SetAlgoLined
        \KwIn{Iterations $K$, sample size $n$, inner update number $M$, policy learning rate $\alpha_\theta$, value learning rate $\alpha_\phi$, trade-off parameter $\lambda$, importance sampling ratio clip $\zeta$; other parameters used by PPO-clip, e.g., PPO-clip range, GAE lambda}
        \textbf{Initialize:}  policy $\pi_\theta$, value function $V_\phi$\\
        \For{$k$ in $1:K$}{
            \tcp{\textcolor{blue}{Sample trajectories}}
            $\{\tau_i\}_{i=1}^n \leftarrow$ run\_episodes($\pi_\theta,n$)
            
            Compute trajectory return $R_{\tau_i}$ and reward-to-go $\{R_{\tau_i,t}\}_{t=0}^{T-1}$ for each $\tau_i$ 
            
            Compute advantages $A(s_{i,t},a_{i,t})$ for each state-action based on current $V_\phi$
            
            \For{$m$ in $1:M$}{
            \tcp{\textcolor{blue}{PPO-clip} }
            Compute ppo\_grad via PPO-clip
            
            \tcp{\textcolor{blue}{Variability PG}}
            Compute importance sampling ratio $\rho_i$ for each $\tau_i$, $\rho_i=\min(\rho_i,\zeta)$
            
            Split $\{\tau_i\}_{i=1}^n$ for double sampling if required
            
            Compute variability\_grad via corresponding equation with $\rho_i$
            
            \tcp{\textcolor{blue}{Update policy}}
            $\theta \leftarrow \theta + \alpha_\theta~ (\mathrm{ppo\_grad} - \lambda~ \mathrm{variability\_grad})$
            
            \tcp{\textcolor{blue}{Update value function}}
            \For{$\mathrm{all~mini\_batch~of~} s_{i,t}$ in ${\{\tau_i\}}_{i=1}^n$}{
            $\phi\leftarrow \phi - \alpha_\phi \nabla_\phi \mathbb{E}[(V_\phi(s_{i,t})-R_{\tau_i,t})^2]$
            }
            }        
        }
\label{algo:ppo_variability}
\end{algorithm}

\textit{Remark.} IS ratio should be applied for each expectation calculation. For instance, in Eq.~\ref{eq:sv-item}, there is an expectation inside another expectation. We should estimate both of them using IS. For Mean-Median Deviation, Inter-Quantile Range, and CVaR Deviation, the quantile estimation should also consider the IS weight, see Appendix~\ref{app:quantile_is}.

\section{Experiments}

We compare the policy gradients using different measures of variability mentioned above in several domains where risk-aversion can be clearly verified to investigate their practical performances of finding risk-averse policies. We use lower tail CVaR Deviation and downside Semi\_Variance and Semi\_STD in our experiments to avoid low returns. For methods requiring upper bound $b$, we use the maximum of $\{R_{\tau_i}\}_{i=1}^n$ in practice. Please refer to Sec.~\ref{sec:exp-detail} in Appendix for any missing implementation details.

\subsection{Tabular Case: Maze}

\begin{wrapfigure}{r}{0.26\textwidth}
\vspace{-0.3in}
  \begin{center}
    \includegraphics[width=0.26\textwidth]{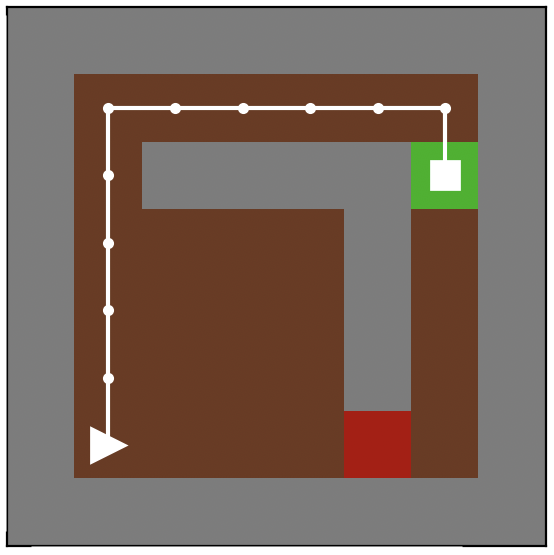}
  \end{center}
  \vspace{-0.1in}
  \caption{A modified Maze. Red state returns an uncertain reward (details in text).}
  \label{fig:risk-maze}
  \vspace{-0.15in}
\end{wrapfigure}

We begin the evaluation in a simple tabular case. This domain is modified from \cite{greenberg2022efficient} as shown in Fig.~\ref{fig:risk-maze}. The original maze is asymmetric with two openings to reach the top path (in contrast to a single opening for the bottom path). In addition, paths via the top tend to be longer than paths via the bottom. We modified the maze to be more symmetric to reduce preferences arising from certain exploration strategies that might be biased towards shorter paths or greater openings, which may confound risk-aversion. Starting from the bottom left corner, the agent aims to reach the green goal state. The gray color corresponds to the walls. Every movement before reaching the goal receives a reward of $-1$, except for the red state, whose reward is random with mean $-1$. Thus, the shortest path going through the red state toward the goal is the optimal risk-neutral path, while the longer path (shown in white color) is risk-averse, though its expected return is slightly lower. We investigate four types of reward distribution for the red state, i.e., Gaussian, Pareto, Uniform, and a handcrafted distribution to simulate outliers. The details of the distribution are described in Appendix~\ref{app:maze}. We add a sketch of the reward distribution to each plot for illustration. The maximum episode length is $100$. Agents collect $n=50$ episodes before updating the policy, without IS. The learning algorithm of agents is Algo.~\ref{algo:rf_variability}. 

\begin{figure}[H]
    \begin{center}
        \includegraphics[width=0.75\textwidth]{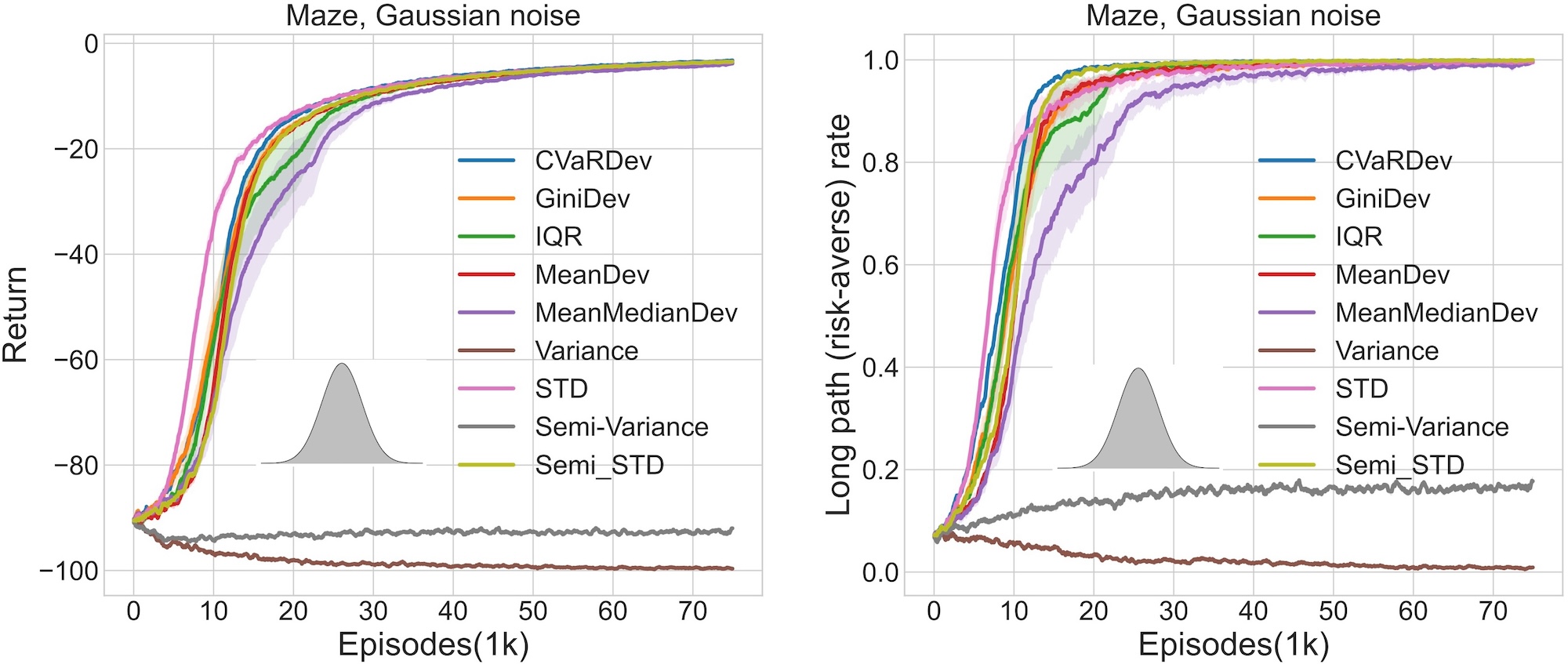}
    \end{center}
    \caption{The reward distribution of the red state is Gaussian. (a) Return and (b) Risk-averse (long path) rate of each algorithm v.s. training episodes in Maze. Curves are averaged over 10
seeds with shaded regions indicating standard errors.}
    \label{fig:maze-gaussian}
\vspace{-13pt}
\end{figure}

\begin{figure}[H]
    \begin{center}
        \includegraphics[width=0.75\textwidth]{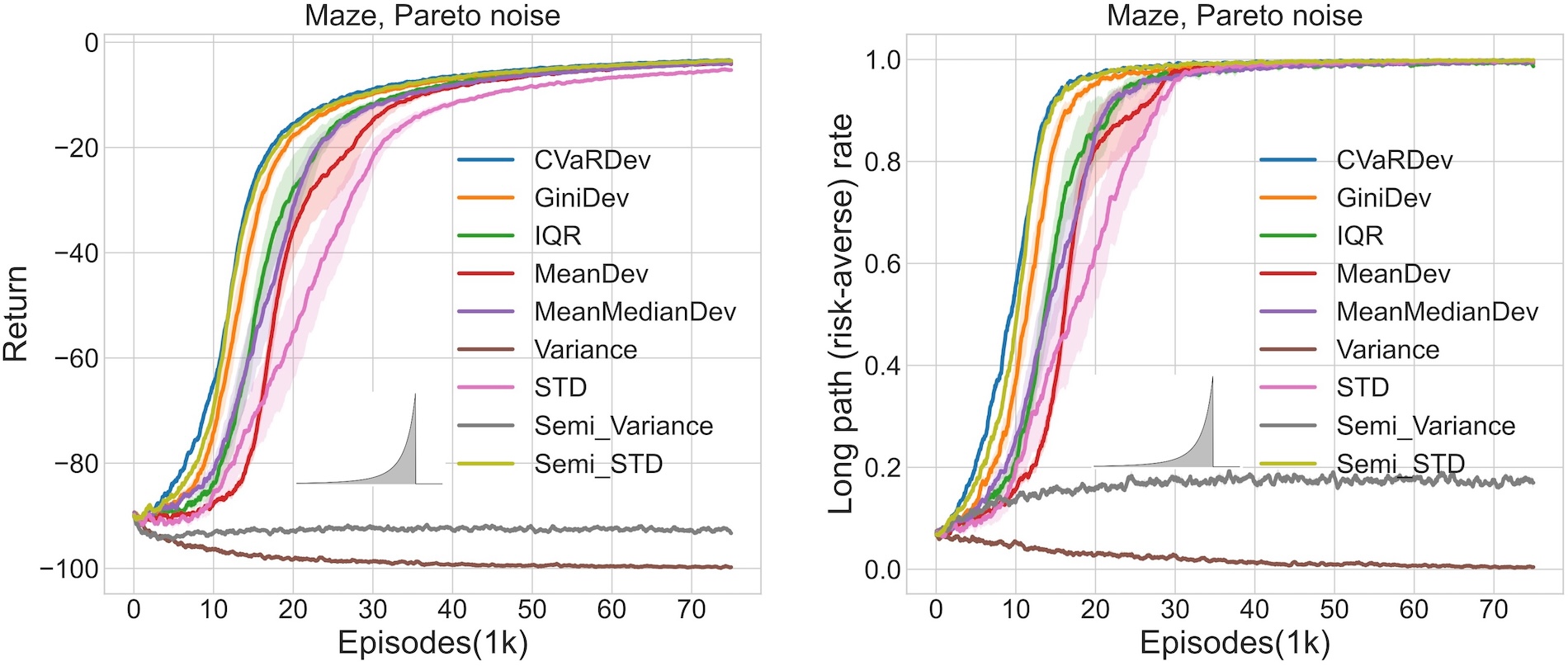}
    \end{center}
    \caption{The reward distribution of the red state is Pareto. (a) Return and (b) Risk-averse (long path) rate of each algorithm v.s. training episodes in Maze. Curves are averaged over 10
seeds with shaded regions indicating standard errors.}
    \label{fig:maze-pareto}
\vspace{-10pt}
\end{figure}

\begin{figure}[H]
    \begin{center}
        \includegraphics[width=0.75\textwidth]{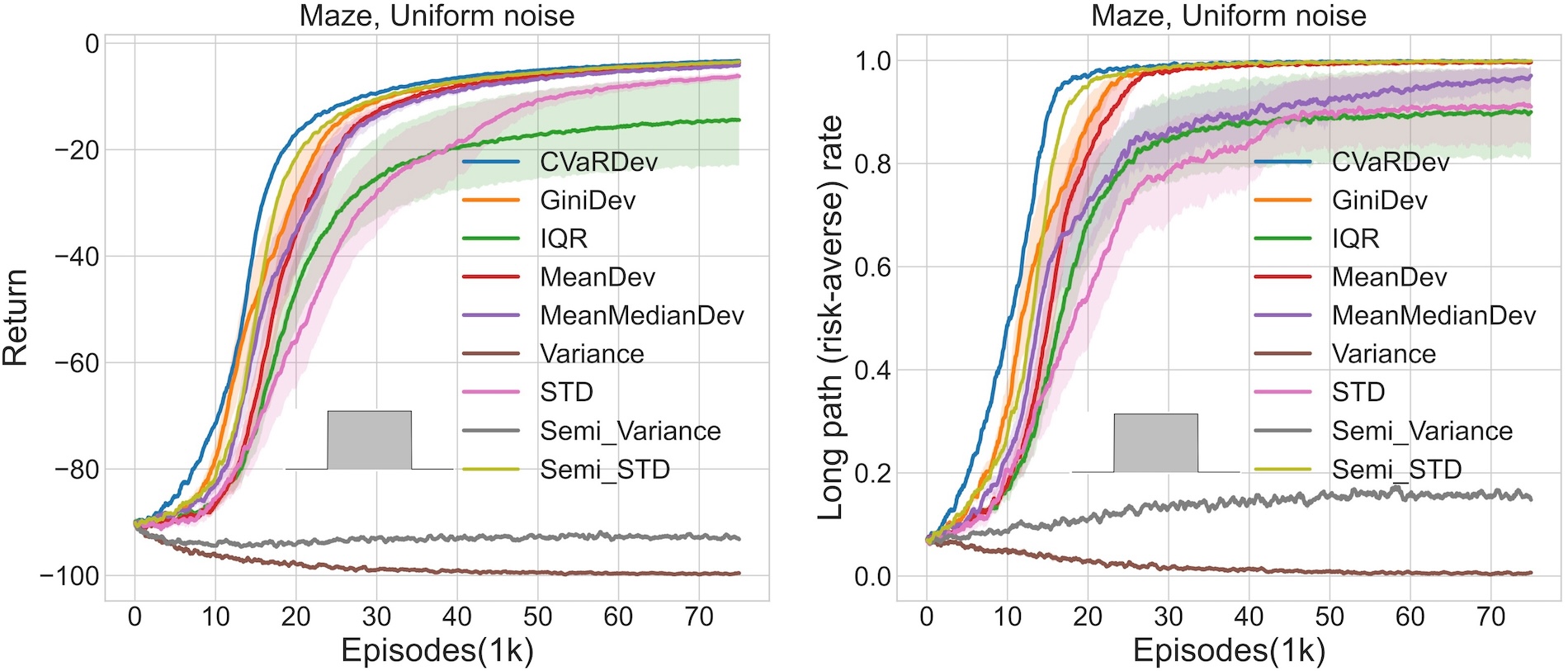}
    \end{center}
    \caption{The reward distribution of the red state is Uniform. (a) Return and (b) Risk-averse (long path) rate of each algorithm v.s. training episodes in Maze. Curves are averaged over 10
seeds with shaded regions indicating standard errors.}
    \label{fig:maze-uniform}
\vspace{-15pt}
\end{figure}

\begin{figure}[H]
    \begin{center}
        \includegraphics[width=0.8\textwidth]{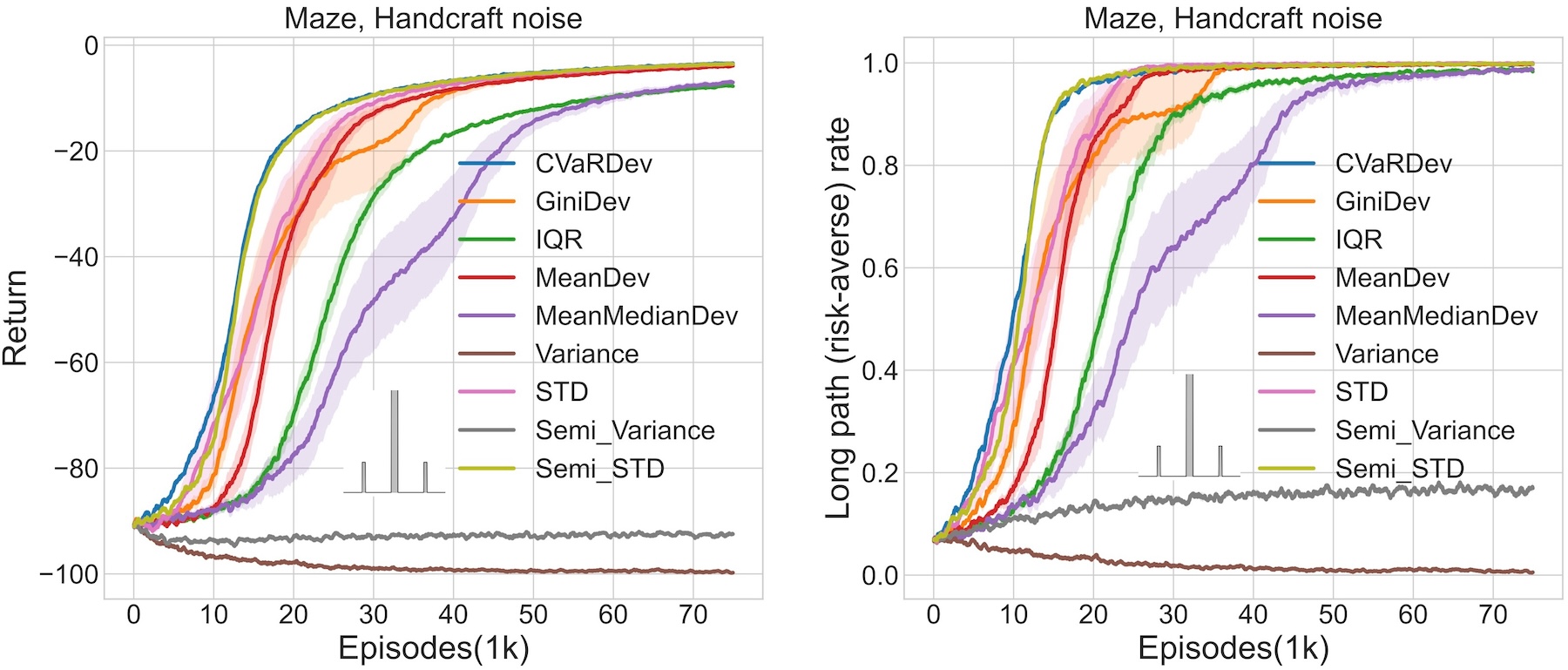}
    \end{center}
    \caption{The reward distribution of red state is a handcraft distribution. (a) Return and (b) Risk-averse (long path) rate of each algorithm v.s. training episodes in Maze. Curves are averaged over 10 seeds with shaded regions indicating standard errors.}
    \label{fig:maze-outlier}
\end{figure}

\begin{figure}[t]
    \begin{center}
        \includegraphics[width=0.99\textwidth]{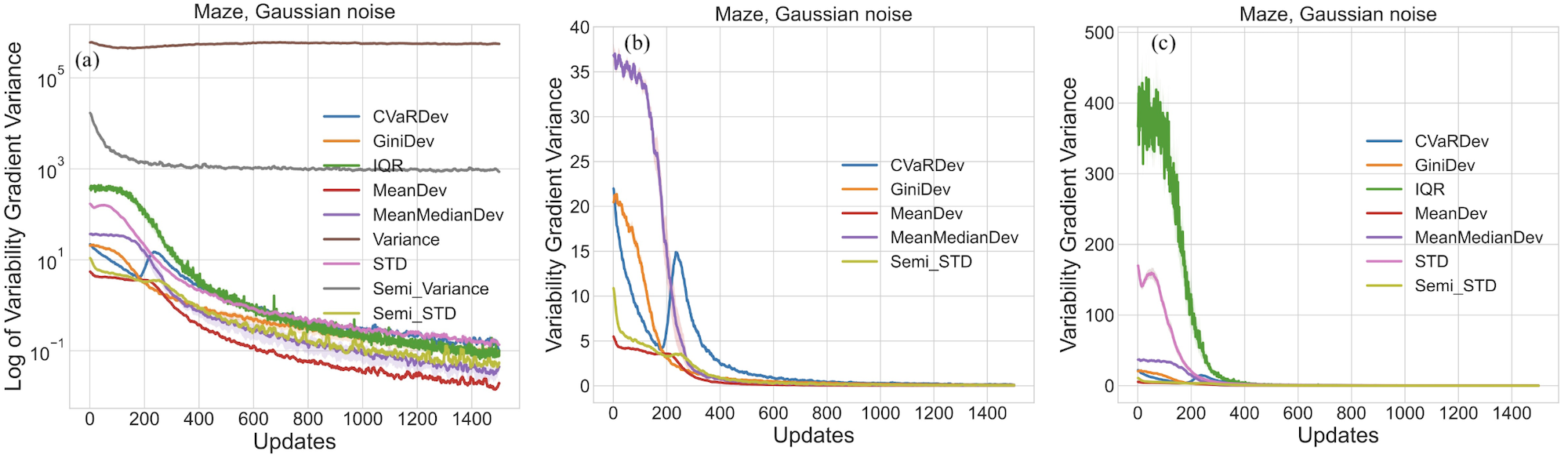}
    \end{center}
    \caption{(a) Logarithm of the gradient variance of different metrics. (b), (c) The gradient variance without using logarithmic scale. (c) adds IQR and STD to (b). Curves are averaged over 10 seeds with shaded regions indicating standard errors.}
    \label{fig:maze-grad-var}
\end{figure}

The return and risk-aversion curves of using each metric under different reward distributions are shown in Fig.~\ref{fig:maze-gaussian}, \ref{fig:maze-pareto}, \ref{fig:maze-uniform}, and \ref{fig:maze-outlier}. For all cases, selecting appropriate hyperparameters for Variance and Semi\_Variance is challenging, as the quadratic term in their gradients leads to unstable updates. To demonstrate this instability, we investigate the gradient variance of variability metrics when the reward noise is Gaussian, as shown in Fig.~\ref{fig:maze-grad-var}. Compared with other metrics, Variance and Semi\_Variance exhibit significantly higher gradient variance during training, making it difficult to stabilize policy updates. We also observe that IQR and STD show a relatively higher gradient variance at the initial stage of training (Fig.~\ref{fig:maze-grad-var} (c)), while the gradient variance of Mean Deviation and Semi\_STD is low (Fig.~\ref{fig:maze-grad-var} (b)).

Apart from Variance and Semi\_Variance, all other metrics successfully identify risk-averse policies. Overall, CVaR Deviation, Gini Deviation, Mean Deviation, and Semi\_STD demonstrate relatively stable risk-averse performance across all four cases, with CVaR Deviation learning slightly faster. However, the performance of the Inter-Quantile Range and Mean-Median Deviation is affected in scenarios where the reward distributions are Uniform or our handcrafted distribution. This is because the policy gradient of Inter-Quantile Range only considers samples between $1-\alpha$ and $\alpha$ quantiles, potentially overlooking deviations outside this range. For instance, we put the outliers below the 0.05 quantile and above the 0.95 quantile in our handcrafted distribution (outside the inter-quantile range), which slows down its learning process. In the Mean-Median Deviation gradient (Eq.~\ref{eq:mmd_grad_est}), the indicator function can cause a trajectory to be selected twice to calculate different parts of the gradient if its return equals the median, and these two parts happen to cancel out each other. This is particularly likely when the return distribution is discrete and highly concentrated, as in the case of our handcrafted distributions, explaining the decline in its performance. STD performs better than Variance as it scales the gradient of variance by a factor, which helps to stabilize (a similar phenomenon occurs with Semi\_STD). However, STD performs well only when the reward distribution is Gaussian or close to Gaussian, such as our handcrafted distribution. Its performance downgrades when the reward distribution is far from Gaussian, as in the case of Pareto and Uniform distributions.

\begin{figure}[t]
    \begin{center}
        \includegraphics[width=0.75\textwidth]{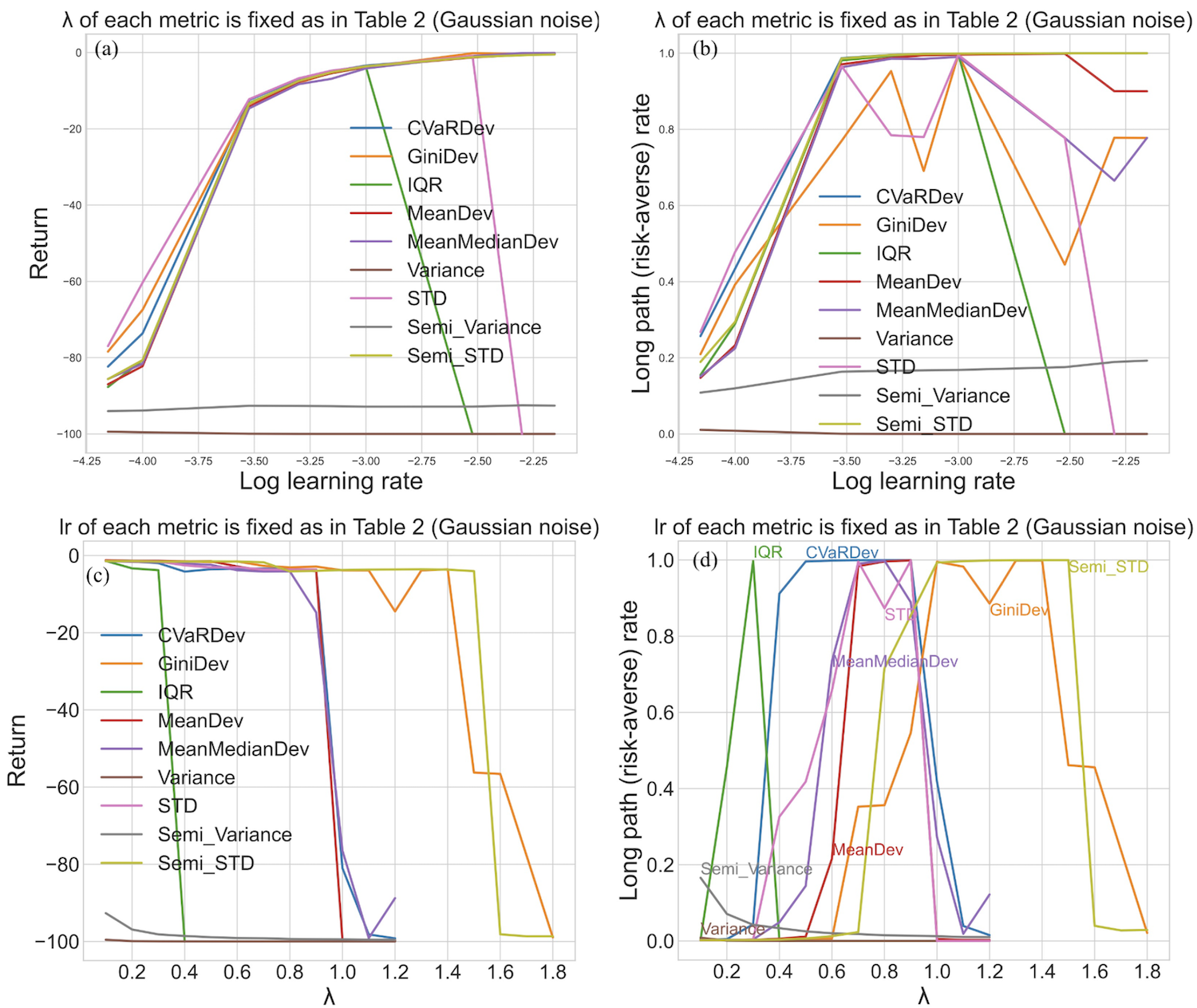}
    \end{center}
    \caption{Sensitivity to hyperparameters when reward noise is Gaussian. (a),(b) Return and risk-averse rate when fixing $\lambda$ as in Table 2. (c),(d) Return and risk-averse rate when fixing learning rate as in Table 2. Curves are averaged over 10 seeds.}
    \label{fig:maze-sensi}
\end{figure}

We further examine the sensitivity of each metric to hyperparameters using Gaussian noise as an example. The two primary hyperparameters are the learning rate and the trade-off parameter $\lambda$. When varying the learning rate, we fix $\lambda$ for each metric as specified in Table~\ref{tab:maze_param}, and vice versa. We report the average return and risk-averse rate achieved in the last 100 training steps, as shown in Fig.~\ref{fig:maze-sensi}. The sensitivity to learning rate is pretty similar among different metrics (except Variance and Semi\_Variance), with CVaR Deviation, Mean Deviation, and Semi\_STD exhibiting a broader range. For sensitivity to $\lambda$, CVaR Deviation and Semi\_STD demonstrate a wider suitable range, followed by Gini Deviation. Notably, inter-Quantile Range is extremely sensitive to $\lambda$ in this scenario.

Next, we evaluate these metrics in more complex domains using deep learning.

\subsection{Discrete Control: LunarLander}

This domain is taken from OpenAI Gym Box2D environments~\citep{brockman2016openai}. We refer readers to its official documents for the full description. The goal of the agent is to land on the ground without crashing. We split the ground into two parts by the middle line of the landing pad, as shown in Fig.~\ref{fig:lunarlander_env} in the Appendix. If landing on the right, an additional noisy reward sampled from $\mathcal{N}(0,1)$ times 100 is given. For lower return variability, a risk-averse agent should learn to land on the left as much as possible. The maximum episode length is 500. Agents collect $n=30$ episodes before updating the policy, without IS. The learning algorithm of agents is Algo.~\ref{algo:rf_variability}.

\begin{figure}[H]
    \begin{center}
        \includegraphics[width=0.75\textwidth]{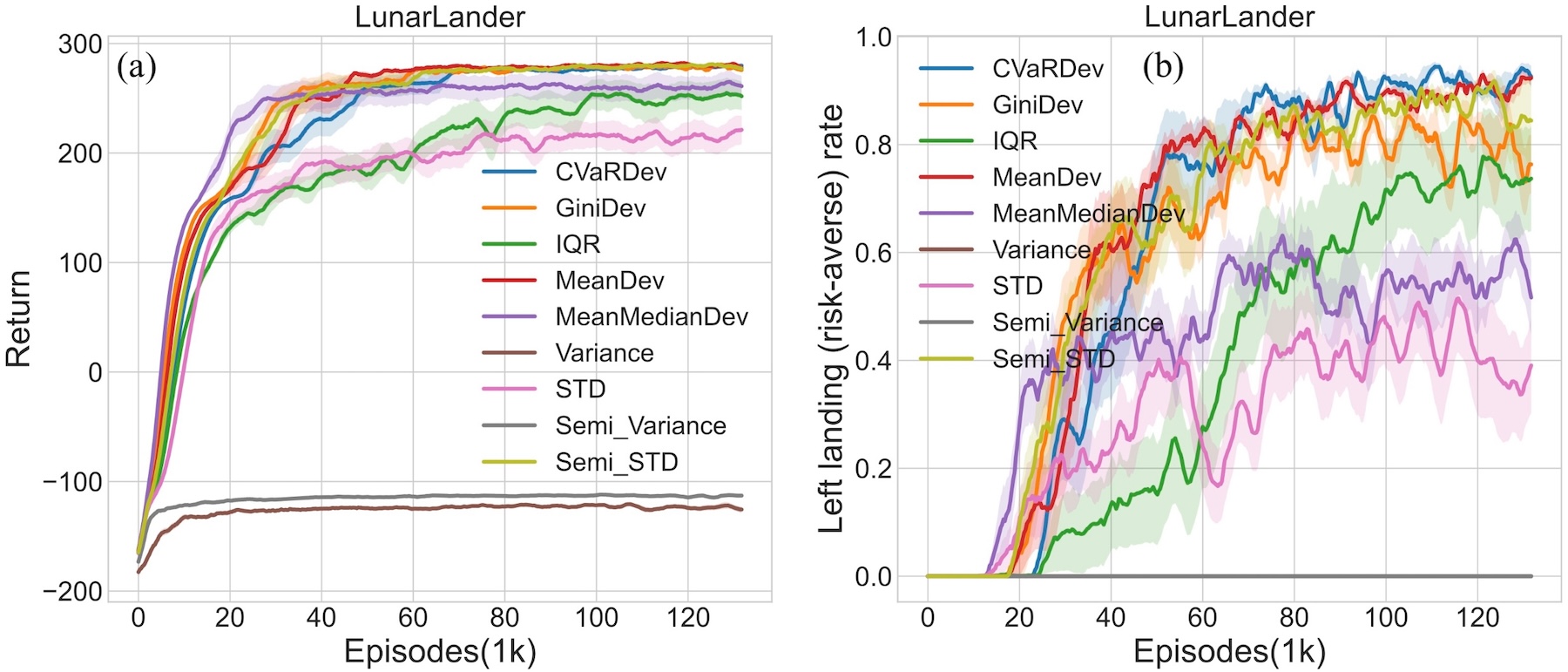}
    \end{center}
    \caption{(a) Policy return and (b) Risk-averse (left-landing) rate v.s. training episodes in LunarLander. Curves are averaged over 10 seeds with shaded regions indicating standard errors.}
    \label{fig:ll_ret_rate}
\vspace{-5pt}
\end{figure}

We present the return and left-landing rates for different metrics in Fig.~\ref{fig:ll_ret_rate}. CVaR Deviation, Gini Deviation, Mean Deviation, and Semi\_STD successfully learn risk-averse policies, with left-landing rates exceeding $80\%$. Notably, CVaR Deviation and Mean Deviation achieve slightly higher risk-averse rates compared to Gini Deviation and Semi\_STD. Inter-Quantile Range yields a relatively lower return and risk-averse rate than these four metrics. In contrast, Mean-Median Deviation and STD fail to learn a clear risk-averse policy, with risk-averse rates below 0.6.

\subsection{Continuous Control: Mujoco}

\begin{figure}[t]
    \begin{center}
        \includegraphics[width=0.75\textwidth]{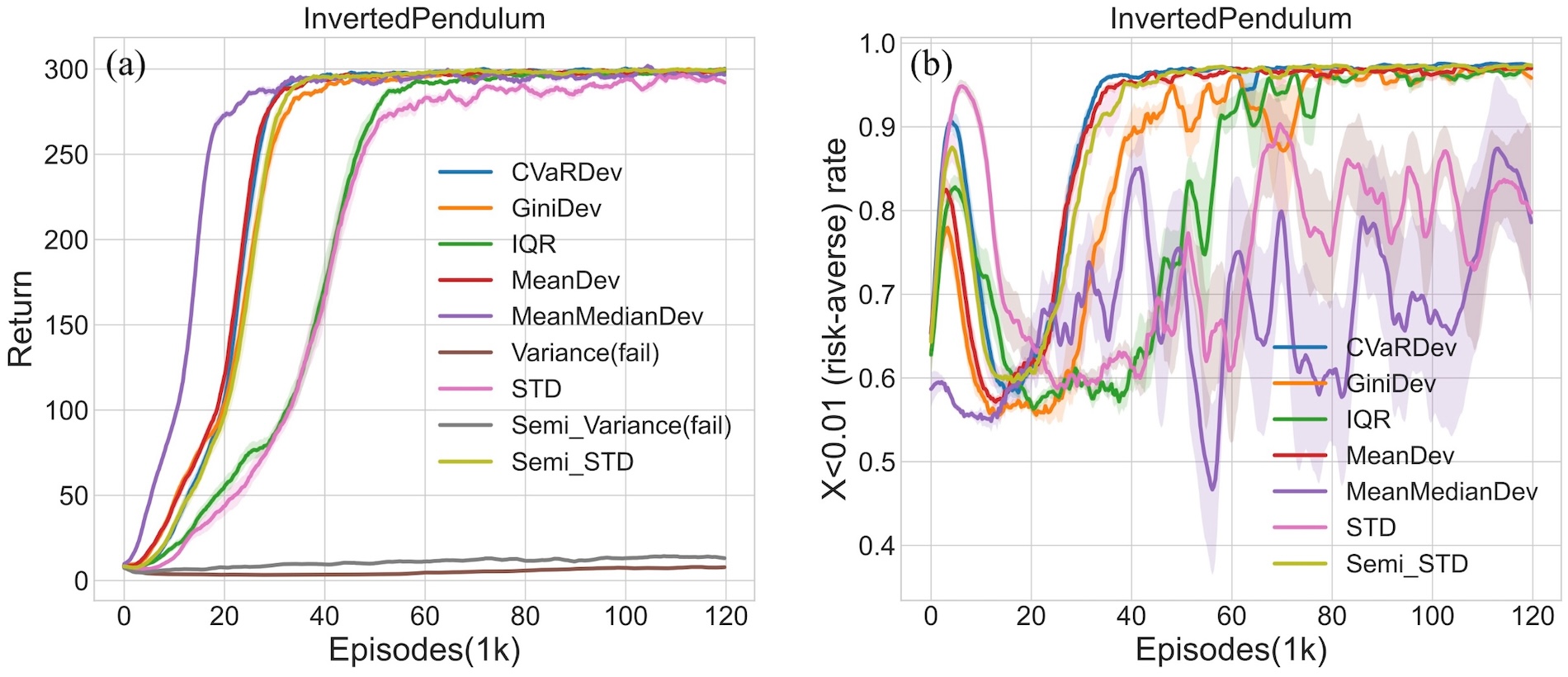}
    \end{center}
    \caption{(a) Policy return and (b) Risk-averse (X$<$0.01) rate v.s. training episodes in InvertedPendulum. Curves are averaged over 10 seeds with shaded regions indicating standard errors.}
    \label{fig:Ivp_ret_rate}
\end{figure}

\begin{figure}[t]
    \begin{center}
        \includegraphics[width=0.8\textwidth]{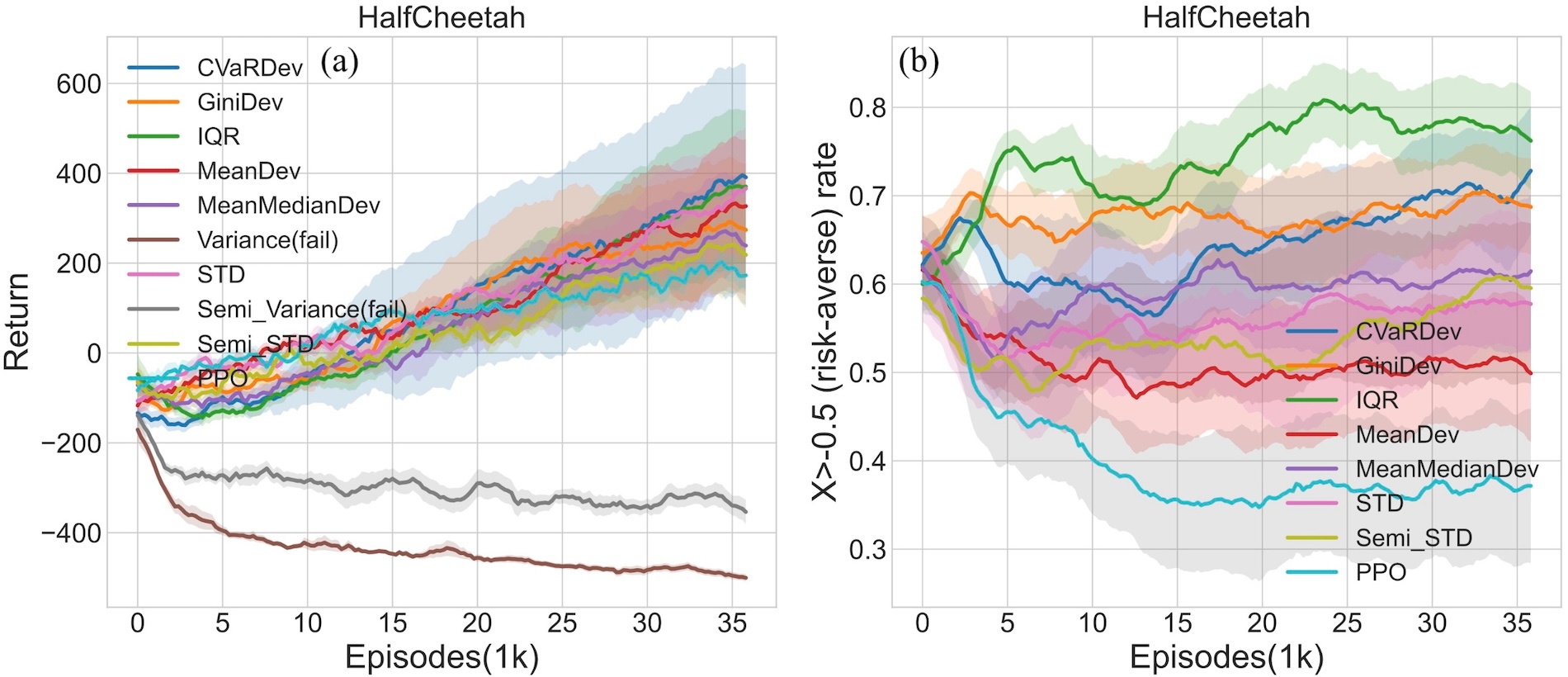}
    \end{center}
    \caption{(a) Policy return and (b) Risk-averse (X$>$-0.5) rate v.s. training episodes in HalfCheetah. Curves are averaged over 10 seeds with shaded regions indicating standard errors.}
    \label{fig:hc_ret_rate}
\end{figure}

Mujoco~\citep{todorov2012mujoco} is a collection of robotics environments with continuous states and actions in OpenAI Gym~\citep{brockman2016openai}. Here, we select two domains, namely InvertedPendulum, and HalfCheetah. Inspired by \cite{liu2023benchmarking}, we define the risky region
based on the X-position. Specifically, if X-position $>0.01$ in InvertedPendulum, X-position $<-0.5$ in HalfCheetah, a zero-mean Gaussian noise is added to the reward ($\mathcal{N}(0,1)\times 10$ in InvertedPendulum, and $\mathcal{N}(0,1)\times 50$ in HalfCheetah). Location information is appended to the observation such that the agent is aware of its current location. To reduce the return variability, a risk-averse agent should reduce the time it visits the noisy region in an episode.  In InvertedPendulum, the maximum episode length is 300. Agents collect $n=30$ episodes before updating the policy, without IS. The learning algorithm of agents is Algo.~\ref{algo:rf_variability}. In HalfCheetah, the maximum episode length is 500. Agents collect $n=12$ episodes before updating the policy. The learning algorithm of agents is Algo~\ref{algo:ppo_variability}. 

To avoid confusion, we did not include curves of Variance and Semi\_Variance in Fig.~\ref{fig:Ivp_ret_rate} (b) and Fig.~\ref{fig:hc_ret_rate} (b) since they fail to learn a reasonable policy.

In InvertedPendulum (Fig.~\ref{fig:Ivp_ret_rate}), CVaR Deviation, Gini Deviation, Mean Deviation, and Semi\_STD achieve comparable returns, with Gini Deviation exhibiting slightly slower progress in risk-aversion. While Mean-Median Deviation achieves higher returns more quickly, it fails to learn a clear risk-averse policy. STD does not perform as well as the others in learning a sufficiently risk-averse policy with a fluctuant risk-averse rate. Inter-Quantile Range eventually demonstrates risk-aversion but converges more slowly compared to the others.

In HalfCheetah (Fig.~\ref{fig:hc_ret_rate}), we include PPO as a risk-neutral baseline, trained using Algo.~\ref{algo:ppo_variability} without the variability gradient. Except for Variance and Semi\_Variance, using other metrics helps to achieve higher returns than PPO, indicating that penalizing return variability enhances policy learning in this case. The returns achieved by using different metrics do not show a significant difference, with CVaR Deviation achieving the highest mean return. However, in terms of risk-averse rate, while all metrics (except Variance and Semi\_Variance) achieve higher risk-aversion than PPO, Inter-Quantile Range performs the best, followed by Gini Deviation and CVaR Deviation.

\textbf{Summary.} Our results reveal several key findings. 
\begin{enumerate}
  \item  Variance-based metrics (Variance, Semi\_Variance), although common, exhibit high gradient variance due to the quadratic term and are difficult to apply in practice. They often fail to effectively learn either risk-averse or risk-neutral policies.
  \item STD-based metrics (STD, Semi\_STD) perform significantly better than variance-based metrics by scaling the gradient. However, STD shows inconsistencies when balancing the tradeoff between high returns and low variability. In some cases, such as LunarLander, STD-based methods fail to achieve either high returns or effective risk-averse behavior. In contrast, Semi\_STD demonstrates smaller gradient variance and more robust performance than STD.
  \item Among the metrics evaluated, CVaR Deviation and Gini Deviation consistently delivered robust and reliable performance across all tested domains and noise scenarios. These metrics not only learned risk-averse policies quickly but also achieved competitive mean returns compared with other variability-based methods.
  \item Mean Deviation, an underexplored metric, and Semi\_STD also exhibited strong performance across different noise settings and environments (except for the HalfCheetah domain). They also demonstrated a smaller gradient variance compared with others.
  \item Though Mean-Median Deviation may be a more robust metric to quantify the risk than Mean Deviation since median is more robust to outliers and skewed distributions, optimizing it via gradient-based approach fails to achieve a clear risk-averse policy in several domains, e.g., LunarLander, InvertedPendulum. 
  \item Although we used a simple KDE method for density estimation in the Inter-Quantile Range gradient, it performed well in several scenarios. It even outperformed STD and Mean-Median Deviation in terms of risk-aversion in LunarLander. However, KDE introduces high variance to the gradient, making it sensitive to hyperparameters. Future research could explore techniques for variance reduction to improve its stability.
\end{enumerate}

\section{Discussions}
In this work, we conducted a comprehensive review of nine measures of variability and their utility in RARL policy gradient algorithms, focusing on their role in penalizing the variability of returns. While many of these metrics are well-documented in economics, finance, or statistics literature, their use in RL remains underexplored. Specifically, four out of the nine metrics have not been previously applied to risk-averse policy gradient methods. We derived the policy gradient formulas for all these metrics and analyzed their corresponding gradient estimation properties. Notably, only Mean-Median Deviation, Inter-Quantile Range, and CVaR Deviation produce biased gradient estimates, while the remaining metrics yield unbiased gradients. For gradients with unintuitive unbiased properties, we provided theoretical proofs to validate their correctness.

We implemented these measures of variability within a mean-variability risk-averse framework using the well-known REINFORCE and popular PPO algorithm. Our empirical study systematically evaluated the trade-off between maximizing expected return and minimizing risk as defined by each metric. To clearly demonstrate the effectiveness of risk-averse policies, we designed synthetic environments or thoughtfully modified existing well-known environments where risk-averse behaviors can be easily identified. This allowed us to observe and analyze the practical utility of these algorithms in risk-aware decision-making. 

The results of our experiment demonstrate the robustness of CVaR Deviation, Gini Deviation, Semi\_STD, and underexplored Mean Deviation across different randomness and evaluation domains, which offers practical insights for researchers and practitioners on the selection of appropriate measures of variability for real-world RL applications where risk-averse behavior is desired. Additionally, our findings lay a foundation for further research into designing and innovating new risk metrics and risk-averse algorithmic frameworks.

\textbf{Limitations and future work.} In terms of theory, we do not provide unbiased policy gradient derivations for certain metrics, e.g., Mean-Median Deviation, Inter-Quantile Range, and CVaR Deviation. This omission is due to the empirical estimation of quantiles.
Additionally, we do not provide a theoretical proof of convergence for the proposed methods. This is largely because there is currently no generalized proof framework that covers those policy gradient methods with penalties by using all those measures of variability used in this work. Based on our assessment, rigorously addressing convergence properties for these variants would require a dedicated study or multiple follow-up works.

In terms of experiments, while our empirical evaluations are thorough in terms of noise settings and types of environments---including both discrete and continuous control tasks, as well as synthetic and more challenging robotic environments---we do not cover all popular RL benchmarks. This is primarily due to the non-trivial challenge of modifying existing environments to clearly identify and observe risk-averse policy behavior. Developing risk-aware environments with reproducible and meaningful risk trade-offs is itself a complex research problem. For example, in certain MuJoCo domains, introducing similar noise conditions led to failures in both risk-neutral and risk-averse learning, making it difficult to isolate and analyze the causes of these failures.

Lastly, although our work comprehensively explores various measures of variability and their application to policy gradient-based RARL algorithms, we have not yet investigated other potential risk metrics that may also lead to risk-averse behavior. Exploring these alternative metrics and their integration into policy gradient frameworks is a promising direction for future research.

These limitations motivate several immediate future efforts. First, we plan to derive unbiased policy gradient estimators or estimators with improved variance-bias tradeoffs for the metrics that currently lack them, such as Mean-Median Deviation, Inter-Quantile Range, and CVaR Deviation. Second, we aim to address the theoretical gap by developing a generalized framework to prove the convergence of policy gradient methods with various measures of variability. Establishing such a framework would provide theoretical guarantees to enhance the reliability of RARL algorithms and potentially offer valuable insights into how hyperparameters, gradient bias or variance, or noise settings affect sample efficiency. Finally, it is also important to study how to improve the sample efficiency of policy gradient methods when using different measures of variability. RARL algorithms are known to suffer from low sample efficiency, which is a major scientific hurdle limiting their deployment in real-world applications. For example, it might be promising to combine risk-neutral and risk-averse approaches using a mixture parameterization \citep{luo2024simple}, where the algorithm dynamically balances between optimizing for expected return and minimizing risk. This could significantly enhance the practical utility of risk-averse RL algorithms by reducing their sample complexity while maintaining robust risk-aware behavior.

Addressing these challenges will provide both theoretical insight and practical advances, ultimately improving the applicability of RARL in high stake real-world domains.

\section*{Acknowledgements}
Resources used in this work were provided, in part, by the Province of Ontario, the Government of Canada through CIFAR, companies sponsoring the Vector Institute
(\url{https://vectorinstitute.ai/partners/}), the Natural Sciences and Engineering Council of Canada, and the Digital Research Alliance of Canada (alliancecan.ca). Yudong Luo is also supported by a Waterloo AI Institute Graduate Scholarship.

\newpage
\onecolumn

\section{Proof of Propositions and Theorems}
\label{app:theorem}
\subsection{Proof of Propositions}
\subsubsection{Mean Deviation}
\label{sec:mdgrad-proof}
\mdgrad*
\noindent\textit{Proof.} Following assumptions~\ref{assump:1}, \ref{assump:2}, \ref{assump:3}, we have
\begin{equation}
    \mathrm{MD}[X] = \int_{-b}^b |x-\mathbb{E}[X]| f_X(x;\theta) dx ,
\end{equation} 
whose gradient w.r.t $\theta$ is
\label{eq:md_x}
\begin{align}
    \nabla_\theta \mathrm{MD}[X]&=\int^b_{-b}|x-\mathbb{E}[X]|\nabla_\theta f_X(x;\theta) dx + \int^b_{-b}f_X(x;\theta) \nabla_\theta |x-\mathbb{E}[X]|dx \\
    &=\int^b_{-b}|x-\mathbb{E}[X]|f_X(x;\theta) \nabla_\theta \log f_X(x;\theta) dx - \int_{-b}^b f(x;\theta) \mathrm{sgn}(x-\mathbb{E}[X])\nabla_\theta \mathbb{E}[X] dx \\
    &=\mathbb{E}_{x\sim X}\Big[|x-\mathbb{E}[X]| \nabla_\theta \log f_X(x;\theta)\Big] - \mathbb{E}_{x\sim X}\Big[\mathrm{sgn}(x-\mathbb{E}[X])\Big]\nabla_\theta\mathbb{E}[X].~~\square
\end{align}

\subsubsection{Mean-Median Deviation}
\label{sec:mmdgrad-proof}
\mmdgrad*
\noindent\textit{Proof.} Using the gradient of quantile in Eq.~\ref{eq:quantile_grad} (recall that under our assumptions, $q_\alpha(X;\theta)= F^{-1}_X(\alpha)$), the gradient of $\mathrm{MMD}[X]$ is
\begin{equation}
\begin{aligned}
    \nabla_\theta \mathrm{MMD}[X] = &-\int_{\frac{1}{2}}^1\int_{-b}^{q_\alpha(X;\theta)} \nabla_\theta f_X(x;\theta) dx \frac{1}{f_X(q_\alpha(X;\theta);\theta)} d\alpha \\
    &+\int_0^{\frac{1}{2}}\int_{-b}^{q_\alpha(X;\theta)} \nabla_\theta f_X(x;\theta) dx \frac{1}{f_X(q_\alpha(X;\theta);\theta)} d\alpha .
\end{aligned}
\end{equation}
Switching the integral order, we get
\begin{equation}
\begin{aligned}
    \nabla_\theta \mathrm{MMD}[X] = &-\int_{-b}^{q_{\frac{1}{2}}(X;\theta)} \int_{\frac{1}{2}}^1 \nabla_\theta f_X(x;\theta) \frac{1}{f_X(F^{-1}(\alpha);\theta)} d\alpha dx \\
    &-\int_{q_{\frac{1}{2}}(X;\theta)}^b \int_{F_X(x)}^1 \nabla_\theta f_X(x;\theta) \frac{1}{f_X(F^{-1}(\alpha);\theta)} d\alpha dx \\
    &+\int_{-b}^{q_{\frac{1}{2}}(X;\theta)} \int_{F_X(x)}^{\frac{1}{2}} \nabla_\theta f_X(x;\theta) \frac{1}{f_X(F^{-1}(\alpha);\theta)} d\alpha dx .
\end{aligned}
\end{equation}
Similar to Eq.~\ref{eq:change_variable}, we further change the inner integral from $d\alpha$ to $d F_X(t)$ with $t=F^{-1}_X(\alpha)$.
\begin{align}
    \nabla_\theta \mathrm{MMD}[X] = &-\int_{-b}^{q_{\frac{1}{2}}(X;\theta)} \nabla_\theta f_X(x;\theta)\int_{q_{\frac{1}{2}}(X;\theta)}^b  \frac{1}{f_X(t;\theta)} d F_X(t) dx \\
    &-\int_{q_{\frac{1}{2}}(X;\theta)}^b \nabla_\theta f_X(x;\theta)\int_{x}^b  \frac{1}{f_X(t;\theta)} d F_X(t) dx \\
    &+\int_{-b}^{q_{\frac{1}{2}}(X;\theta)} \nabla_\theta f_X(x;\theta)\int_{x}^{q_{\frac{1}{2}}(X;\theta)}  \frac{1}{f_X(t;\theta)} d F_X(t) dx \\
    = & - \int_{-b}^{q_{\frac{1}{2}}(X;\theta)} \nabla_\theta f_X(x;\theta) \big(b-q_{\frac{1}{2}}(X;\theta)\big) dx - \int_{q_{\frac{1}{2}}(X;\theta)}^b \nabla_\theta f_X(x;\theta) (b-x) dx\\
    &+\int_{-b}^{q_{\frac{1}{2}}(X;\theta)}\nabla_\theta f_X(x;\theta) (q_{\frac{1}{2}}(X;\theta)-x) dx .
\end{align}
Applying $\nabla_\theta \log(x)=\frac{1}{x}\nabla_\theta x$ to $f_X(x;\theta)$, we have 
\begin{align}
    \nabla_\theta\mathrm{MMD}[X]=&~~\frac{1}{2}\int_{-b}^{q_{\frac12(X;\theta)}} \frac{1}{1/2} f_X(x;\theta) \nabla_\theta \log f_X(x;\theta)(2q_{\frac12}(X;\theta)-x-b) dx\\
    &-\frac{1}{2}\int_{q_{\frac12}(X;\theta)}^b \frac{1}{1/2} f_X(x;\theta) \nabla_\theta \log f_X(x;\theta)(b-x)dx\\
    =&~~\frac12 \mathbb{E}\Big[\nabla_\theta\log f_X(X;\theta)(2q_{\frac12}(X;\theta)-X-b)| X\leq q_{\frac12}(X;\theta)\Big]\\
    &-\frac12 \mathbb{E}\Big[\nabla_\theta \log f_X(X;\theta)(b-X)|X\geq q_{\frac12}(X;\theta)\Big].  ~~\square
\end{align}

\subsubsection{Inter-Quantile Range}
\label{sec:iqrgrad-proof}
\iqrgrad*
\noindent\textit{Proof.} From the definition of IQR (Eq.\ref{eq:iqr_def}), we have
\begin{equation}
    \nabla_\theta \mathrm{IQR}_\alpha[X] = \nabla_\theta F^{-1}_X(\alpha) - \nabla_\theta F^{-1}_X(1-\alpha).
\end{equation}
Using the gradient of quantile in Eq.~\ref{eq:quantile_grad}, we have
\label{eq:iqr_grad}
\begin{align}
    \nabla_\theta \mathrm{IQR}_\alpha[X]=&-\frac{1}{f_X(q_\alpha(X;\theta);\theta)} \int_{-b}^{q_\alpha(X;\theta)} \nabla_\theta f_X(x;\theta) dx\\
    &+ \frac{1}{f_X(q_{1-\alpha}(X;\theta);\theta)}\int_{-b}^{q_{1-\alpha(X;\theta)}}\nabla_\theta f_X(x;\theta) dx \\
    =&-\frac{\alpha}{f_X(q_\alpha(X;\theta);\theta)} \int_{-b}^{q_\alpha(X;\theta)} \frac{1}{\alpha}f_X(x;\theta) \nabla_\theta \log f_X(x;\theta) dx\\ 
    &+ \frac{1-\alpha}{f_X(q_{1-\alpha}(X;\theta);\theta)}\int_{-b}^{q_{1-\alpha(X;\theta)} }\frac{1}{1-\alpha}f_X(x;\theta)\nabla_\theta f_X(x;\theta)dx\\
    = &-\frac{\alpha}{f_X(q_\alpha(X;\theta);\theta)} \mathbb{E}\Big[\nabla_\theta \log f_X(x;\theta)|X\leq q_\alpha(X;\theta)\Big] \\
    &+\frac{1-\alpha}{f_X(q_{1-\alpha}(X;\theta);\theta)} \mathbb{E}\Big[\nabla_\theta \log f_X(x;\theta)|X\leq q_{1-\alpha}(X;\theta)\Big]. ~~\square
\end{align}

\subsubsection{CVaR Deviation}
\label{sec:cdgrad-proof}
\cdgrad*
\noindent\textit{Proof.} Since the gradient of the mean term is easy to get in RL, we only need to compute the gradient of the CVaR term. The gradient of CVaR is previously given by \cite{tamar2015optimizing}. We still provide a computation here for completeness. 

Under assumptions~\ref{assump:1}, \ref{assump:2}, \ref{assump:3}, we have
\begin{equation}
    \mathrm{CVaR}^\lor_\alpha(X)=\mathbb{E}[X|X\leq q_\alpha(X;\theta)]=\frac{1}{\alpha}\int_{-b}^{q_\alpha(X;\theta)} f_X(x;\theta) x dx .
\end{equation}
Taking the derivative and using the Leibniz rule we obtain
\begin{equation}
\label{eq:cvar_grad_im}
    \nabla_\theta \mathrm{CVaR}^\lor_\alpha(X) = \frac{1}{\alpha} \int_{-b}^{q_\alpha(X;\theta)} \nabla_\theta f_X(x;\theta) x dx + \frac{1}{\alpha} \nabla_\theta q_\alpha(X;\theta) f_X(q_\alpha(X;\theta);\theta) q_\alpha(X;\theta) .
\end{equation}
Plugging the gradient of quantile (Eq.~\ref{eq:quantile_grad})  into Eq.~\ref{eq:cvar_grad_im}, and rearranging, we obtain
\begin{align}
    \nabla_\theta \mathrm{CVaR}_\alpha ^\lor(X) &= \frac{1}{\alpha}\int_{-b}^{q_{\alpha}(X;\theta)}\nabla_\theta f_X(x;\theta)(x-q_\alpha(X;\theta))dx\\
    &= \int_{-b}^{q_\alpha(X;\theta)} \frac{1}{\alpha} f_X(x;\theta) \nabla_\theta \log f_X(x;\theta) (x-q_\alpha(X;\theta)) dx \\
    &=\mathbb{E}\Big[\nabla_\theta \log f_X(X;\theta) (X-q_\alpha(X;\theta))| X\leq q_\alpha(X;\theta)\Big]. ~~\square
\end{align}

\subsubsection{Semi\_Variance}
\label{sec:svargrad-proof}
\svargrad*
\noindent\textit{Proof.} Under assumptions~\ref{assump:1}, \ref{assump:2}, \ref{assump:3}, Semi\_Variance can be expressed alternatively as
\begin{equation}
    {\mathrm{S}}\mathbb{V}[X] = \int_{-b}^{\mathbb{E}[X]}(x-\mathbb{E}[X])^2 f_X(x;\theta) dx , 
\end{equation}
whose gradient is (by applying the Leibniz rule)
\begin{align}
    \nabla_\theta {\mathrm{S}}\mathbb{V}[X] =& \int_{-b}^{\mathbb{E}[X]} \nabla_\theta \Big[(x-\mathbb{E}[X])^2 f_X(x;\theta)\Big] dx\\
    =&\int_{-b}^{\mathbb{E}[X]}(x-\mathbb{E}[X])^2 \nabla_\theta f_X(x;\theta) dx + \int_{-b}^{\mathbb{E}[X]} 2(\mathbb{E}[X]-x) \nabla_\theta \mathbb{E}[X] f_X(x;\theta) dx\\
    =& \int_{-b}^{\mathbb{E}[X]}(x-\mathbb{E}[X])^2 \nabla_\theta \log f_X(x;\theta) \cdot f_X(x;\theta) dx \\
    &+ \nabla_\theta \mathbb{E}[X]\int_{-b}^{\mathbb{E}[X]} 2(\mathbb{E}[X]-x) f_X(x;\theta) dx\\
    =&\mathbb{E}\Big[(X-\mathbb{E}[X])^2 \nabla_\theta\log f_X(X;\theta)\mathbb{I}_{X\leq \mathbb{E}[X]}\Big]\\
    &+ \nabla_\theta \mathbb{E}[X] \cdot\mathbb{E}\Big[2(\mathbb{E}[X]-X)\mathbb{I}_{X\leq \mathbb{E}[X]}\Big]. ~~\square
\end{align}

\subsection{Proof of Theorem 1}
\label{sec:thm1-proof}
\mmdconsist*
\textit{Proof.} There are two terms in the calculation of $\nabla_{\theta}\widehat{\mathrm{MMD}}[X]_{[n]}$ (Eq.~\ref{eq:mmd_grad_est}). We only prove that one of them is consistent and the other bears the similarity. Here, we take the first term as an example.

\begin{equation}
\label{eq:mmd-term1}
    \frac{1}{n}\sum_{i=1}^n(2\hat{q}_{\frac{1}{2}}-x_i-b) \nabla_\theta f_X(x_i;\theta) \mathbb{I}_{x_i\leq \hat{q}_{\frac12}},
\end{equation}
where $\hat{q}_{\frac12}$ is the empirical half quantile. For simplicity, define $h_1(x)=(x+b)\nabla_\theta f_X(x;\theta)$ and $h_2(x)=\nabla_\theta f_X(x;\theta)$. We use the abbreviation $q_{\frac12}$ for the true half quantile $q_{\frac12}(X;\theta)$. Then Eq.~\ref{eq:mmd-term1} equals to
\begin{equation}
\label{eq:mmd-expand}
\begin{aligned}
    &\frac{1}{n}\sum_{i=1}^n\Big(2\hat{q}_{\frac12} h_2(x_i)-h_1(x_i)\Big) \mathbb{I}_{x_i\leq \hat{q}_{\frac12}}\\
    =&\frac{1}{n}\sum_{i=1}^n\Big(2 q_{\frac12} h_2(x_i)-h_1(x_i)\Big)\mathbb{I}_{x_i\leq {q}_{\frac12}}\\
    +& \frac{1}{n}\sum_{i=1}^n\Big(2q_{\frac12}h_2(x_i)-h_1(x_i)\Big)\Big(\mathbb{I}_{x_i\leq \hat{q}_{\frac12}}-\mathbb{I}_{x_i\leq {q}_{\frac12}}\Big) + \frac{1}{n}\sum_{i=1}^n 2(\hat{q}_{\frac12}-q_\frac12) h_2(x_i) \mathbb{I}_{x_i\leq \hat{q}_{\frac12}} .
\end{aligned}
\end{equation}

We show that the two extra terms in the last line of Eq.~\ref{eq:mmd-expand} vanish when $n\rightarrow\infty$.
Denote $2\hat{q}_{\frac{1}{2}}h_2(x_i)-h_1(x_i)$ as $l(x_i,\hat{q}_{\frac12})$. By our assumption, $l(x_i,\hat{q}_{\frac12})$ is bounded. By Holder's inequality,
\begin{equation}
    \Big|\frac{1}{n}\sum_{i=1}^nl(x_i,\hat{q}_{\frac12})\Big(\mathbb{I}_{x_i\leq \hat{q}_{\frac12}}-\mathbb{I}_{x_i\leq {q}_{\frac12}}\Big)\Big|\leq\Big(\frac{1}{n}\sum_{i=1}^n |l(x_i),\hat{q}_{\frac12}|^2\Big)^{\frac12} \Big(\frac{1}{n}\sum_{i=1}^n|\mathbb{I}_{x_i\leq \hat{q}_{\frac12}}-\mathbb{I}_{x_i\leq {q}_{\frac12}}|^2\Big)^{\frac12} .
\end{equation}
Note that
\begin{equation}
\begin{aligned}
    &\frac{1}{n}\sum_{i=1}^n|\mathbb{I}_{x_i\leq \hat{q}_{\frac12}}-\mathbb{I}_{x_i\leq {q}_{\frac12}}|^2 = \frac{1}{n}\sum_{i=1}^n|\mathbb{I}_{x_i\leq \hat{q}_{\frac12}}-\mathbb{I}_{x_i\leq {q}_{\frac12}}|\\
    =&(\mathbb{I}_{\hat{q}_{\frac12}>q_{\frac12}} - \mathbb{I}_{\hat{q}_{\frac12}\leq q_\frac12}) \frac{1}{n}\sum_{i=1}^n(\mathbb{I}_{x_i\leq \hat{q}_{\frac12}}-\mathbb{I}_{x_i\leq {q}_{\frac12}}) .
\end{aligned}
\end{equation}
By Proposition 4.1 of \cite{hong2009simulating}, w.p. 1, $\frac{1}{n}\sum_{i=1}^n(\mathbb{I}_{x_i\leq \hat{q}_{\frac12}}-\mathbb{I}_{x_i\leq {q}_{\frac12}}) \rightarrow 0$ when $n\rightarrow\infty$. Also, with the continuous mapping theory~\citep{durrett2019probability}, we have  $\Big(\frac{1}{n}\sum_{i=1}^n|\mathbb{I}_{x_i\leq \hat{q}_{\frac12}}-\mathbb{I}_{x_i\leq {q}_{\frac12}}|^2\Big)^{\frac12} \rightarrow 0$ when $n\rightarrow\infty$. Therefore, $\frac{1}{n}\sum_{i=1}^nl(x_i,\hat{q}_{\frac12})\Big(\mathbb{I}_{x_i\leq \hat{q}_{\frac12}}-\mathbb{I}_{x_i\leq {q}_{\frac12}}\Big) \rightarrow 0$ when $n\rightarrow \infty$.

It is shown by \cite{david2004order} that the sample quantile is a consistent estimator of the true quantile, i.e., $\hat{q}_{\frac12}-q_{\frac12}\rightarrow 0$ as $n\rightarrow\infty$, thus $\frac{1}{n}\sum_{i=1}^n 2(\hat{q}_{\frac12}-q_\frac12) h_2(x_i) \mathbb{I}_{x_i\leq \hat{q}_{\frac12}}\rightarrow\infty$ when $n\rightarrow \infty$.

Therefore, the two extra terms vanish and 
\begin{equation}
\label{eq:mmd-pf1}
    \frac{1}{n}\sum_{i=1}^n(2\hat{q}_{\frac{1}{2}}-x_i-b) \nabla_\theta f_X(x_i;\theta) \mathbb{I}_{x_i\leq \hat{q}_{\frac12}} \rightarrow \frac{1}{n}\sum_{i=1}^n(2{q}_{\frac{1}{2}}-x_i-b) \nabla_\theta f_X(x_i;\theta) \mathbb{I}_{x_i\leq {q}_{\frac12}} ,
\end{equation}
when $n \rightarrow \infty$.

Similarly, we have
\begin{equation}
\label{eq:mmd-pf2}
    \frac{1}{n}\sum_{i=1}^n(b-x_i)\nabla_\theta \log f_X(x;\theta) \mathbb{I}_{x_i\geq \hat{q}_{\frac12}} \rightarrow \frac{1}{n}\sum_{i=1}^n(b-x_i)\nabla_\theta \log f_X(x;\theta) \mathbb{I}_{x_i\geq {q}_{\frac12}} .
\end{equation}
Combing Eq.~\ref{eq:mmd-pf1} and \ref{eq:mmd-pf2}, we have $\nabla_{\theta}\widehat{\mathrm{MMD}}[X]_{[n]}\rightarrow \nabla_\theta \mathrm{MMD}[X]$ as $n\rightarrow \infty$. $~\square$

\subsection{Proof of Theorem 2}
\label{sec:thm2-proof}
\mmdbound*
\textit{Proof.} Similar to the proof of Theorem 1, we show that the error of the first term of $\nabla_{\theta}\widehat{\mathrm{MMD}}[X]_{[n]}$ is $O(n^{-\frac12})$ and the other bears the similarity.

Use the abbreviation $q_{\frac12}$ for the true half quantile $q_{\frac12}(X;\theta)$. The first term of $\nabla_\theta \mathrm{MMD}[X]$ is
\begin{equation}
\label{eq:mmd1-orig}
\begin{aligned}
    &\frac12 \mathbb{E}\Big[(2q_{\frac12}-X-b)\nabla_\theta \log f_X(X;\theta)|X\leq q_\frac12\Big]\\
    =&\frac12\times \frac{1}{1/2}\mathbb{E}\Big[(2q_{\frac12}-X-b)\nabla_\theta \log f_X(X;\theta) \mathbb{I}_{X\leq q_{\frac12}}\Big] .
\end{aligned}
\end{equation}
The first term of $\nabla_{\theta}\widehat{\mathrm{MMD}}[X]_{[n]}$ is
\begin{equation}
\label{eq:mmd1-split}
\begin{aligned}
    &\frac{1}{2}\times \frac{1}{n/2} \sum_{i=1}^n(2\hat{q}_{\frac{1}{2}}-x_i-b) \nabla_\theta f_X(x_i;\theta) \mathbb{I}_{x_i\leq \hat{q}_{\frac12}}\\
    =&\frac12 \times \frac{1}{n/2}\sum_{i=1}^n(2q_{\frac{1}{2}}-x_i-b) \nabla_\theta f_X(x_i;\theta) \mathbb{I}_{x_i\leq \hat{q}_{\frac12}} + \frac12\times \frac{2}{n/2} \sum_{i=1}^n(\hat{q}_{\frac12}-q_\frac12)\nabla_\theta f_X(x_i;\theta)\mathbb{I}_{x_i\leq \hat{q}_{\frac12}} .
\end{aligned}
\end{equation}

Here, we will use a theory from \cite{hong2009simulating} to assist the proof. Denote $D_1(X;\theta)=(2q_{\frac12}(X;\theta)-X-b)\nabla_\theta \log f_X(X;\theta)$

\begin{theorem}[\cite{hong2009simulating}]
\label{thm:hong}
    Assume assumptions~\ref{assump:1}, \ref{assump:2}, \ref{assump:3}, \ref{assump:4} hold. Let
    \begin{equation}
        \bar{L}_n = \frac{1}{\alpha n}\sum_{i=1}^n D_1(x_i) \mathbb{I}_{x_i\leq \hat{q}_\alpha} ,
    \end{equation}
    where $x_i$ is an i.i.d sample of $X$, and let
    \begin{equation}
        L = \frac{1}{\alpha} \mathbb{E}[D_1(X)\mathbb{I}_{X\leq q_\alpha}] , 
    \end{equation}
    then $\mathbb{E}[\bar{L}_n]-L$ is  $o(n^{-\frac12})$.
\end{theorem}

Following theorem~\ref{thm:hong}, we have that the difference between the expectation of the first term in Eq.~\ref{eq:mmd1-split} and Eq.~\ref{eq:mmd1-orig} is $o(n^{-\frac12})$, i.e., $\mathbb{E}\Big[\frac{1}{n/2}\sum_{i=1}^n(2q_{\frac{1}{2}}-x_i-b) \nabla_\theta f_X(x_i;\theta) \mathbb{I}_{x_i\leq \hat{q}_{\frac12}}\Big]-\frac{1}{1/2}\mathbb{E}\Big[(2q_{\frac12}-X-b)\nabla_\theta \log f_X(X;\theta) \mathbb{I}_{X\leq q_{\frac12}}\Big]$ is $o(n^{-\frac12})$.

Now we analyze the remaining term of Eq.~\ref{eq:mmd1-split}, i.e., 
\begin{equation}
\label{eq:mmd1-split-2}
    \frac12\times \frac{2}{n/2} \sum_{i=1}^n(\hat{q}_{\frac12}-q_\frac12)\nabla_\theta f_X(x_i;\theta)\mathbb{I}_{x_i\leq \hat{q}_{\frac12}} .
\end{equation}
By assumptions \ref{assump:2} and \ref{assump:4}, $|\nabla_\theta\log f_X(X;\theta)|$ is bounded and assume it is bounded by $\bar{h}$. 

Note that $|\frac{1}{n}\sum_{i=1}^n \nabla_\theta f_X(x_i;\theta)\mathbb{I}_{x_i\leq \hat{q}_{\frac12}}|\leq \bar{h}$. Therefore,
\begin{equation}
    \begin{aligned}
    \mathbb{E}\Big[(\hat{q}_{\frac12}-q_{\frac12})\frac{1}{n}\sum_{i=1}^n \nabla_\theta f_X(x_i;\theta)\mathbb{I}_{x_i\leq \hat{q}_{\frac12}}\Big]&\leq \mathbb{E}\Big[|\hat{q}_\frac12-q_\frac12|\times |\frac{1}{n}\sum_{i=1}^n \nabla_\theta f_X(x_i;\theta)\mathbb{I}_{x_i\leq \hat{q}_{\frac12}}|\Big]\\
    &\leq \bar{h}~\mathbb{E}\Big[|\hat{q}_\frac12 - q_\frac12|\Big] .
\end{aligned}
\end{equation}

It is well known that the difference between the empirical quantile and the true quantile is $O(n^{-\frac12})$, e.g., see \cite{lahiri2009berry}. Thus, Eq.~\ref{eq:mmd1-split-2} (which is the second term of the r.h.s. of Eq.~\ref{eq:mmd1-split}) is bounded by $O(n^{-\frac12})$.

Combining the results we have, we can conclude that the error of the expectation of Eq.~\ref{eq:mmd1-split} is $O(n^{-\frac{1}{2}})$.

Similarly, the error of the expectation of the second term in $\nabla_{\theta}\widehat{\mathrm{MMD}}[X]_{[n]}$ is $O(n^{-\frac12})$, which means $\mathbb{E}\big[\nabla_{\theta}\widehat{\mathrm{MMD}}[X]_{[n]}\big] - \nabla_\theta \mathrm{MMD}[X]$ is $O(n^{-\frac{1}{2}})$. $~\square$



\newpage
\section{Justification of Coherent Measures of Variability}
\subsection{Mean Deviation}
\label{sec:md-coherent}
It is obvious that Mean Deviation satisfies (A) Law-invariance and (C1) Standardization.

(C2) Location invariance: $\mathrm{MD}[X-c]=\mathbb{E}[|X-c -\mathbb{E}[X-c]|]=\mathbb{E}[|X-\mathbb{E}[X]|]=\mathrm{MD}[X]$.

(A1) Positive homogeneity: $\mathrm{MD}[cX]=\mathbb{E}[|cX-\mathbb{E}[cX]|]=c\mathbb{E}[|X-\mathbb{E}[X]|]=c \mathrm{MD}[X]$, for $c>0$.

(A2) Sub-additivity: $\mathrm{MD}[X+Y]=\mathbb{E}[|X+Y-\mathbb{E}[X+Y]|]=\mathbb{E}[|X-\mathbb{E}[X] + Y-\mathbb{E}[Y]|]\leq \mathbb{E}[|X-\mathbb{E}[X]|]+\mathbb{E}[|Y-\mathbb{E}[Y]|]=\mathrm{MD}[X] + \mathrm{MD}[Y]$.

Therefore, Mean Deviation is a coherent measure of variability.

\subsection{Mean-Median Deviation}
\label{sec:mmd-coherent}
It is obvious that Mean-Median Deviation satisfies (A) Law-invariance and (C1) Standardization.

(C2) Location invariance: $\mathrm{MMD}[X-c]=\mathbb{E}[|X-c-\mathrm{Median(X-c)}|]=\mathbb{E}[|X-\mathrm{Median}(X)|]=\mathrm{MMD}[X]$.

(A1) Positive homogeneity: $\mathrm{MMD}[cX]=\mathbb{E}[|cX-\mathrm{Median}(cX)|]=c\mathbb{E}[|X-\mathrm{Median}(X)|]=c \mathrm{MMD}[X]$, for $c>0$.

(A2) Sub-additivity: $\mathrm{MMD}[X+Y]=\mathbb{E}[|X+Y-\mathrm{Median}(X+Y)|]\leq\mathbb{E}[|X+Y-(\mathrm{Median}(X)+\mathrm{Median}(Y))|]= \mathbb{E}[|X-\mathrm{Median}(X)+Y-\mathrm{Median}(Y)|]\leq\mathbb{E}[|X-\mathrm{Median}(X)|]+\mathbb{E}[|Y-\mathrm{Median}(Y)|]=\mathrm{MMD}[X] + \mathrm{MMD}[Y]$. The first inequality is due to $\mathrm{Median}(X+Y)=\arg\min_z \mathbb{E}[|X+Y-z|]$.

Therefore, Mean-Median Deviation is a coherent measure of variability.

\subsection{Upper tail CVaR Deviation}
\label{sec:cd-coherent}
Upper tail CVaR is defined as $\mathrm{CVaR}^\land_\alpha(X) = \frac{1}{1-\alpha}\int_{\alpha}^1q_\beta(X) d\beta$, where $q_\beta(X)$ is the quantile.

It is obvious that CVaR Deviation satisfies (A) Law-invariance and (C1) Standardization.

(C2) Location invariance: $\mathrm{CD}[X-c]=\mathrm{CVaR}^\land_\alpha(X -c- \mathbb{E}[X-c])=\mathrm{CVaR}^{\land}_\alpha(X-\mathbb{E}[X])=\mathrm{CD}[X]$.

(A1) Positive homogeneity: $\mathrm{CD}[cX]=\mathrm{CVaR}^{\land}_\alpha(cX-\mathbb{E}[cX])=c\mathrm{CVaR}^{\land}_\alpha(X-\mathbb{E}[X])=c\mathrm{CD}[X]$, for $c>0$.

(A2) Sub-additivity: $\mathrm{CD}[X+Y]=\mathrm{CVaR}^{\land}_\alpha(X+Y-\mathbb{E}[X+Y])=\mathrm{CVaR}^{\land}_\alpha(X-\mathbb{E}[X]+Y-\mathbb{E}[Y])\leq \mathrm{CVaR}^{\land}_\alpha(X-\mathbb{E}[X])+\mathrm{CVaR}^{\land}_\alpha(Y-\mathbb{E}[Y])=\mathrm{CD}[X]+\mathrm{CD}[Y]$. The inequality is due to the sub-additivity of upper tail CVaR.

Therefore, upper tail CVaR Deviation is a coherent measure of variability. Consequently, lower tail CVaR Deviation is not coherent since it fails to be sub-additive.

\subsection{Semi\_STD}
\label{sec:sstd-coherent}
Note that Semi\_STD can be expressed alternatively as $\mathrm{SD}[X]=(\mathbb{E}[(X-\mathbb{E}[X])_{-}^2])^{\frac12}$ for downside or $\mathrm{SD}[X]=(\mathbb{E}[(X-\mathbb{E}[X])_{+}^2])^{\frac12}$ for upside, where $(x)_{-}=-\min(x,0)$, and $(x)_{+}=\max(x,0)$. Note that 
\begin{equation}
\label{eq:xy-ineq}
\begin{aligned}
(x+y)_{-}&\leq (x)_{-} + (y)_{-}\\
(x+y)_{+}&\leq (x)_{+} + (y)_{+}~~\mathrm{for~all}~x,y\in\mathbb{R}.
\end{aligned}
\end{equation}

It is obvious that Semi\_STD satisfies (A) Law-invariance and (C1) Standardization. We show the remaining axioms using downside Semi\_STD as an example.

(C2) Location invariance: $\mathrm{SD}[X-c]=(\mathbb{E}[(X-c-\mathbb{E}[X-c])_{-}^2])^{\frac12}=(\mathbb{E}[(X-\mathbb{E}[X])_{-}^2])^{\frac12}=\mathrm{SD}[X]$.

(A1) Positive homogeneity: $\mathrm{SD}[cX]=(\mathbb{E}[(cX-\mathbb{E}[cX])_{-}^2])^{\frac12}=(\mathbb{E}[c^2(X-\mathbb{E}[X])_{-}^2])^{\frac12}=c(\mathbb{E}[(X-\mathbb{E}[X])_{-}^2])^{\frac12}=c\mathrm{SD}[X]$, for $c>0$.

(A2) Sub-additivity: $\mathrm{SD}[X+Y]=\Big(\mathbb{E}\big[(X+Y-\mathbb{E}[X+Y])_{-}^2\big]\Big)^{\frac12}=\Big(\mathbb{E}\big[(X-\mathbb{E}[X]+Y-\mathbb{E}[X])_{-}^2\big]\Big)^{\frac12}\leq \Big(\mathbb{E}\big[\big((X-\mathbb{E}[X])_{-}+(Y-\mathbb{E}[Y])_{-}\big)^2\big]\Big)^{\frac12}\leq \Big(\mathbb{E}\big[(X-\mathbb{E}[X])_{-}^2\big]\Big)^{\frac12}+\Big(\mathbb{E}\big[(Y-\mathbb{E}[Y])_{-}^2\big]\Big)^{\frac12}=\mathrm{SD}[X]+\mathrm{SD}[Y]$. The first inequality is due to Eq.~\ref{eq:xy-ineq}. The second inequality follows the Minkowski inequality.

The upside Semi\_STD also satisfies the above axioms.
\section{Experiments Details}
\label{sec:exp-detail}
\subsection{Summary of Coherence and PG Properties}
We summarize the coherence of different metrics discussed in the paper, and whether their gradients are unbiased or require double sampling in Table~\ref{tab:var_summary}.
\begin{table}[ht]
\centering
\begin{tabular}{l|c|c|c}
\toprule
         & \multicolumn{1}{c|}{Coherent} & PG unbiased & PG double sampling \\ 
         \midrule
CVaRDev(upper)   &  $\checkmark$   & $\times$ & $\times$  \\ 
GiniDev   &  $\checkmark$  & $\checkmark$ &    $\times$\\ 
IQR       &  $\times$   & $\times$ & $\times$ \\ 
MeanDev   &  $\checkmark$   & $\checkmark$ &    $\checkmark$ \\ 
MeanMedianDev & $\times$ & $\times$ &  $\times$ \\
Variance & $\times$ & $\checkmark$ & $\checkmark$ \\
STD       & $\checkmark$ & $\checkmark$ & $\checkmark$ \\
Semi\_Var & $\times$ &$\checkmark$ & $\checkmark$\\
Semi\_STD & $\checkmark$ & $\checkmark$ & $\checkmark$ \\
\bottomrule
\end{tabular}
\caption{Summary of coherence satisfied by different metrics, and whether their gradients are unbiased or requires double sampling.}
\label{tab:var_summary}
\end{table}

\subsection{The Maze Problem}
\label{app:maze}
The maze is a $6\times 6$ grid. The initial state of the agent is fixed in the bottom left corner. The action space is four (up, down, left, right). The maximum episode length is $100$.

The Gaussian noise is $\mathcal{N}(-1,1)\times 20$. The Pareto noise is $-1 - \big((\mathrm{Pareto}(3,1)-0.5)\times20\big)$. The Uniform noise is $U[-25,23]$. The handcrafted distribution is a mixture of three Uniform distributions, i.e., $U[-2, 0]$, $U[-57, -56]$, and $U[54, 55]$, with probability $0.95, 0.05, 0.05$.

\noindent\textit{Policy function.} The policy is represented as
\begin{equation}
   \pi_\theta(a|s) = \frac{e^{W(s,a)\cdot\theta}}{\sum_b e^{W(s,b)\cdot\theta}} ,
\end{equation}
where $W(s,a)$ is the state-action feature vector, a one-hot encoding in our implementation. Thus, the dimension of $W(s,a)$ is $6\times 6 \times 4$. The derivative of the logarithm is
\begin{equation}
    \nabla_\theta \log \pi_\theta(a|s) = W(s,a)-\mathbb{E}_{b\sim \pi_\theta(\cdot|s)}W(s,b) .
\end{equation}

\noindent\textit{Value function.} The value function of REINFORCE baseline is represented as $V_\upsilon(s)=W(s)\cdot \upsilon$. Similarly, $\zeta(s)$ is a one-hot encoding.

\subsubsection{Learning Parameters}
Discount factor $\gamma=0.999$. Optimizer is stochastic gradient descent. Policy learning rate is chosen from \{1e-3, 7e-4, 5e-4, 3e-4, 1e-4, 7e-5\}. Value learning rate is ten times policy learning rate. Trade-off parameter $\lambda$ is chosen from \{0.1, 0.2, 0.3, 0.4, 0.5, 0.6, 0.7, 0.8, 0.9, 1.0, 1.1, 1.2, 1.3, 1.4, 1.5\}. $\alpha=0.2$ for CVaR Deviation. $\alpha=0.9$ for Inter-Quantile Range. We select a learning rate that is suitable for the majority of methods to ensure that their policy update speeds are approximately on the same scale for a fair comparison. If it is too large for some method, we choose the closest suitable one.

Policy learning rates and $\lambda$s of different methods in four noise cases are summarized in Table~\ref{tab:maze_param}.

\begin{table}[ht]
\centering
\begin{tabular}{l|cc|cc|cc|cc}
\toprule
\multirow{2}{*}{} & \multicolumn{2}{c|}{Gaussian} & \multicolumn{2}{c|}{Pareto} & \multicolumn{2}{c|}{Uniform} & \multicolumn{2}{c}{Handcraft} \\ \cline{2-9} 
                              & lr\_policy & $\lambda$ & lr\_policy & $\lambda$ & lr\_policy & $\lambda$ & lr\_policy & $\lambda$\\ \midrule
CVaRDev   & 1e-3    & 0.6   & 1e-3   & 0.6   & 1e-3 & 0.7 & 1e-3 & 0.6\\ 
GiniDev   & 1e-3    & 1.0   & 1e-3   & 1.3   & 1e-4 & 1.4 & 1e-3 & 1.3\\ 
IQR       & 1e-3    & 0.3   & 1e-3   & 0.3   & 7e-4 & 0.3 & 5e-4 & 0.3\\ 
MeanDev   & 1e-3    & 0.8   & 1e-3   & 0.9   & 1e-3 & 0.9 & 1e-3 & 0.8\\
MeanMedianDev & 1e-3 & 0.7   & 1e-3   & 0.8   & 1e-3 & 0.8 & 7e-4 & 0.8\\
Variance  & 1e-4    & 0.1   & 1e-4   & 0.1   & 1e-4 & 0.1 & 1e-4 & 0.1 \\
STD       & 1e-3    & 0.7   & 7e-4   & 1.0   & 7e-4 & 1.0 & 1e-3 & 1.0\\
Semi\_Var  & 5e-4   & 0.1   & 5e-4   & 0.1   & 5e-4 & 0.1 & 5e-4 & 0.1\\
Semi\_STD  & 1e-3   & 1.2   & 1e-3   & 1.2   & 1e-3 & 1.3 & 1e-3 & 1.2\\
\bottomrule
\end{tabular}
\caption{Policy learning rates and $\lambda$s of different methods in four noise cases in Maze}
\label{tab:maze_param}
\vspace{-10pt}
\end{table}

\begin{figure}[t]
    \begin{center}
        \includegraphics[width=0.4\textwidth]{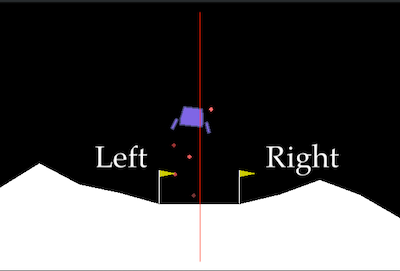}
    \end{center}
    \caption{Split the ground of LunarLander into left and right parts by the middle (red) line. If landing on the right, an additional reward sampled from $\mathcal{N}(0,1)$ times $100$ is given.}
    \label{fig:lunarlander_env}
\vspace{-10pt}
\end{figure}

\subsection{LunarLander Discrete}
The goal is to land the agent on the ground without crashing. The state dimension is $8$. The action dimension is $4$. Detailed reward information is available on this webpage~\footnote{\url{https://www.gymlibrary.dev/environments/box2d/lunar\_lander/}}. Here, we split the ground into left and right parts by the middle line of the landing pad, as shown in Fig.~\ref{fig:lunarlander_env}. If the agent lands on the right part of the ground, an additional noisy reward signal $\mathcal{N}(0, 1)\times 100$ is given. The maximum episode length is $500$.

\noindent\textit{Policy function.} The policy is a categorical distribution, modeled by a neural network. Hidden layer: 2. Hidden size: 128. Activation: ReLU. Softmax function is applied to the output to generate categorical probabilities.

\noindent\textit{Value function.} Hidden layer: 2. Hidden size: 128. Activation: ReLU.

\subsubsection{Learning Parameters}
Discount factor $\gamma=0.999$. Optimizer is Adam. Policy learning rate is chosen from \{1e-3, 7e-4, 5e-4, 3e-4\}. Value learning rate is ten times policy learning rate. Trade-off parameter $\lambda$ is chosen from \{0.2, 0.4, 0.6, 0.8\}. $\alpha=0.2$ for CVaR Deviation. $\alpha=0.9$ for Inter-Quantile Range. We select a learning rate that is suitable for the majority of methods to ensure that their policy update speeds are approximately on the same scale for a fair comparison. If it is too large for some method, we choose the closest suitable one.

Policy learning rates and $\lambda$s of different methods are summarized in Table~\ref{tab:ll_param}.
\begin{table}[ht]
\centering
\begin{tabular}{l|cc|c|l|cc}
\toprule
         & lr\_policy & $\lambda$ & & &lr\_policy & $\lambda$ \\ 
         \midrule
CVaRDev   & 7e-4    & 0.4 & & Variance & 3e-4   & 0.4   \\ 
GiniDev   & 7e-4    & 0.6 & & STD & 7e-4   & 0.4   \\ 
IQR       & 7e-4    & 0.4 & & Semi\_Var & 7e-4   & 0.4   \\ 
MeanDev   & 7e-4    & 0.6 &  & Semi\_STD & 7e-4   & 0.6   \\ 
MeanMedianDev & 7e-4 & 0.6 & &  &   &  \\
\bottomrule
\end{tabular}
\caption{Policy learning rates and $\lambda$s of different methods in LunarLander}
\label{tab:ll_param}
\vspace{-10pt}
\end{table}

\subsection{InvertedPendulum}
The description of Mujoco environments can be found on this webpage~\footnote{\url{https://www.gymlibrary.dev/environments/mujoco/}}. The goal is to balance an inverted pendulum on a cart. The state dimension is 4 (X-position is already contained in the observation). The action dimension is 1. Per step reward is 1. If the agent reaches the region X-position $>$0.01, an additional noisy reward sampled from $\mathcal{N}(0,1)$ times 10 is given. The game ends if the angle between the pendulum and the cart is greater than 0.2 radian or a maximum episode length of 300 is reached.

\noindent\textit{Policy function.} The policy is a normal distribution modeled by a neural network. Hidden layer: 2. Hidden size: 128. Activation: ReLU. Tanh is applied to the last layer. The logarithm of the standard deviation is an independent trainable parameter.

\noindent\textit{Value function.} Hidden layer: 2. Hidden size: 128. Activation: ReLU.

\subsubsection{Learning Parameters}
Discount factor $\gamma=0.999$. Optimizer is Adam. Policy learning rate is chosen from \{3e-4, 1e-4, 7e-5\}. Value learning rate is ten times policy learning rate. Trade-off parameter $\lambda$ is chosen from \{0.2, 0.4, 0.6, 0.8\}. $\alpha=0.2$ for CVaR Deviation. $\alpha=0.9$ for Inter-Quantile Range. We select a learning rate that is suitable for the majority of methods to ensure that their policy update speeds are approximately on the same scale for fair comparison.

Policy learning rates and $\lambda$s of different methods are summarized in Table~\ref{tab:lvp_param}.

\begin{table}[t]
\centering
\begin{tabular}{l|cc|c|l|cc}
\toprule
         & lr\_policy & $\lambda$ & & &lr\_policy & $\lambda$ \\ 
         \midrule
CVaRDev   & 1e-4    & 0.4 & & Variance & 1e-4   & 0.2   \\ 
GiniDev   & 1e-4    & 0.6 & & STD & 1e-4   & 0.6   \\ 
IQR       & 1e-4    & 0.2 & & Semi\_Var & 1e-4   & 0.4   \\ 
MeanDev   & 1e-4    & 0.6 &  & Semi\_STD & 1e-4   & 0.8   \\ 
MeanMedianDev & 1e-4 & 0.6 & &  &   &  \\
\bottomrule
\end{tabular}
\caption{Policy learning rates and $\lambda$s of different methods in InvertedPendulum}
\label{tab:lvp_param}
\vspace{-10pt}
\end{table}

\subsection{HalfCheetah}
The agent controls a robot with two legs. The state dimension is 18 (add X-position). The action dimension is 6. One part of the reward is determined by the distance covered between the current and the previous time step. Originally, it is positive only when the agent moves in the forward (right) direction. To encourage the agent to freely move forward (left) and backward (right), we modify this part of the reward to make it positive as long as the agent is moving far from the origin. If X-position $<$-0.5, an additional noisy reward sampled from $\mathcal{N}(0,1)$ times 50 is given. The game ends when a maximum episode length of 500 is reached.

\noindent\textit{Policy function.} Hidden size: 256. Others are the same as in the case of InvertedPendulum.

\noindent\textit{Value function.} Hidden size: 256. Others are the same as in the case of InvertedPendulum.

\subsubsection{Learning Parameters}
Discount factor $\gamma=0.99$. Optimizer is Adam. Policy learning rate is chosen from \{3e-4, 1e-4, 7e-5\}. Value learning rate is the same as policy learning rate. Trade-off parameter $\lambda$ is chosen from \{0.02, 0.05, 0.07, 0.1, 0.15\}. PPO-clip range is 0.2. GAE lambda is 0.95. $\alpha=0.2$ for CVaR Deviation. $\alpha=0.9$ for Inter-Quantile Range. We select a learning rate that is suitable for the majority of methods to ensure that their policy update speeds are approximately on the same scale for fair comparison. If it is too large for some method, we choose the closest suitable one.

Policy learning rates and $\lambda$s of different methods are summarized in Table~\ref{tab:hc_param}.

\begin{table}[t]
\centering
\begin{tabular}{l|cl|c|l|cl}
\toprule
         & lr\_policy & $\lambda$ & & &lr\_policy & $\lambda$ \\ 
         \midrule
CVaRDev   & 3e-4    & 0.1 & & Variance & 1e-4   & 0.1   \\ 
GiniDev   & 3e-4    & 0.1 & & STD & 3e-4   & 0.05   \\ 
IQR       & 3e-4    & 0.05 & & Semi\_Var & 1e-4   & 0.1   \\ 
MeanDev   & 3e-4    & 0.05 &  & Semi\_STD & 3e-4   & 0.05   \\ 
MeanMedianDev & 3e-4 & 0.05 & & PPO & 3e-4  & - \\
\bottomrule
\end{tabular}
\caption{Policy learning rates and $\lambda$s of different methods in HalfCheetah}
\label{tab:hc_param}
\vspace{-10pt}
\end{table}

\subsubsection{Quantile Estimation using Importance Sampling}
\label{app:quantile_is}
Empirical quantile estimation is required by gradients of Mean-Median Deviation, Inter-Quanitle Range, and CVaR Deviation. When IS weights are $1$, we use \texttt{numpy.quantile(x, alpha)} to get the empirical quantile, where \texttt{x} is the sample set and $\texttt{alpha}$ is the quantile level. When IS weights are not $1$, we normalize the weights and use $\texttt{numpy.interp(alpha, w, x)}$ to get the empirical quantile, where \texttt{alpha} is the quantile level, \texttt{w} is the normalized IS weight, and \texttt{x} is the sample set.

\section{Per-step Reward Variance Methods}
\label{sec:per-step-r-var}
A recent perspective uses per-step reward variance $\mathbb{V}[R]$ as a proxy for $\mathbb{V}[G_0]$. 

The probability mass function of $R$ is $\mathrm{Pr}(R=x) = \sum_{s,a} d_\pi(s,a)\mathbb{I}_{r(s,a)=x}$,
where $d_\pi(s,a) = (1-\gamma) \sum_{t=0}^\infty \gamma^t \mathrm{Pr}(S_t=s, A_t=a|\pi,P)$ is the normalized discounted state-action distribution. Then we have $\mathbb{E}[R]=(1-\gamma)\mathbb{E}[G_0]$ and $\mathbb{V}[G_0]  \leq \frac{\mathbb{V}[R]}{(1-\gamma)^2}$ (see Lemma 1 of \citet{bisi2020risk}). Thus, \citet{bisi2020risk} considers the following objective
\begin{align}
\label{eq:mv_r}
    \max_\pi \mathbb{E}[R] - \lambda \mathbb{V}[R]
    =\mathbb{E}[R - \lambda(R-\mathbb{E}[R])^2].
\end{align}
This objective can be cast as a risk-neutral problem in the original MDP, but with a new reward function $\hat{r}(s,a)=r(s,a)-\lambda\big(r(s,a)-(1-\gamma)\mathbb{E}[G_0]\big)^2$. However, this $\hat{r}(s,a)$ is nonstationary (policy-dependent) due to the occurrence of $\mathbb{E}[G_0]$, so standard risk-neutral RL algorithms cannot be directly applied. Instead, \cite{bisi2020risk} uses trust region optimization~\citep{schulman2015trust} to solve.

\citet{zhang2021mean} introduces Fenchel duality to Equation~\ref{eq:mv_r}. The transformed objective is $\max_\pi \mathbb{E}[R] - \lambda\mathbb{E}[R^2]+\lambda \max_y(2\mathbb{E}[R] y - y^2)$, which equals to 
\begin{equation}
    \max_{\pi,y}J_\lambda(\pi,y)=\sum_{s,a}d_\pi(s,a)\big(r(s,a)-\lambda r(s,a)^2 + 2\lambda r(s,a)y\big) - \lambda y^2.
\end{equation}
The dual variable $y$ and policy $\pi$ are updated iteratively. In each inner loop $k$, $y$ has analytical solution $y_{k+1}=\sum_{s,a}d_{\pi_k}(s,a)r(s,a)=(1-\gamma)\mathbb{E}_{\pi_k}[G_0]$ since it is quadratic for $y$. After $y$ is updated, learning $\pi$ is a risk-neutral problem in the original MDP, but with a new modified reward $\hat{r}(s,a)=r(s,a)-\lambda r(s,a)^2 + 2\lambda r(s,a)y_{k+1}$. Since $\hat{r}(s,a)$ is now stationary, any risk-neutral RL algorithms can be applied for policy updating.

\paragraph{Discussion on Per-step Reward Variance.} Though per-step reward methods have the merit of transferring a risk-averse problem to a risk-neutral one, their limitations are twofold. 1) $\mathbb{V}[R]$ is not an appropriate surrogate for $\mathbb{V}[G_0]$ due to fundamentally different implications. Consider a simple example.  Suppose the policy, the transition dynamics and the rewards are all deterministic, then $\mathbb{V}[G_0]=0$ while $\mathbb{V}[R]$ is usually nonzero unless all the per-step rewards are equal. In this case, shifting a specific step reward by a constant will not affect $\mathbb{V}[G_0]$ and should not alter the optimal risk-averse policy. However, such a shift can lead to a big difference for $\mathbb{V}[R]$ and may result in an invalid policy as we demonstrated in a later example. 2) Reward modification hinders policy learning. Since the reward modifications in~\citet{bisi2020risk} and \citet{zhang2021mean}  share the same issue, here we take \cite{zhang2021mean}  as an example ($\hat{r}(s,a)=r(s,a)-\lambda r(s,a)^2 + 2\lambda r(s,a)y_{k+1}$). This modification is likely to convert a positive reward $r(s,a)$ to a much smaller or even negative value due to the square term, i.e. $-\lambda r(s,a)^2$. In addition, at the beginning of the learning phase, when the policy performance is not good, $y$ is likely to be negative in some environments (since $y$ relates to $\mathbb{E}[G_0]$). Thus, the third term $2\lambda r(s,a)y$ decreases the reward value even more. This prevents the agent from visiting the good (i.e., rewarding) state even if that state does not contribute any risk. These two limitations raise a great challenge to subtly choose the value for $\lambda$ and design the reward for the environment. We refer readers to \cite{luo2023alternative} for more empirical evidence and results.

\setlength{\bibsep}{6pt plus 0.3ex}
\bibliographystyle{apalike}
\bibliography{reference}

\end{document}